\documentclass[sigconf, nonacm]{acmart}

\newcommand\vldbdoi{XX.XX/XXX.XX}
\newcommand\vldbpages{XXX-XXX}
\newcommand\vldbvolume{16}
\newcommand\vldbissue{9}
\newcommand\vldbyear{2023}
\newcommand\vldbauthors{\authors}
\newcommand\vldbtitle{\shorttitle} 
\newcommand\vldbavailabilityurl{https://github.com/zheng-yp/DecoupledDGNN}
\newcommand\vldbpagestyle{plain}

\usepackage{amsmath,amsfonts,dsfont,multirow,multicol,epsfig,url,array,makecell,balance,color,epstopdf}
\usepackage{algorithmic}
\usepackage[ruled, vlined, linesnumbered]{algorithm2e}
\usepackage{hhline}
\usepackage{array}
\usepackage{enumerate}
\usepackage{enumitem}
\usepackage{subfigure}
\usepackage{booktabs}
\usepackage{xcolor,colortbl}
\usepackage{bm}
\usepackage{lipsum}
\usepackage{tabularx}
\usepackage{diagbox}
\usepackage[normalem]{ulem}
\useunder{\uline}{\ul}{}
\setlist[itemize]{leftmargin=*}
\newtheorem{definition}{Definition}

\newtheorem{theorem}{Theorem}

\def\epi{\bm{\hat{\pi}}}
\def\r{\bm{r}}

\def\vpi{\bm{\pi}}

\def\x{\bm{x}}

\def\r{\bm{r}}
\def\Z{\mathbf{Z}}

\def\I{\mathbf{I}}
\def\P{\mathbf{P}}
\def\A{\mathbf{A}}
\def\D{\mathbf{D}}

\def\Q{\mathbf{Q}}

\def\U{\mathbf{U}}
\def\Y{\mathbf{Y}}
\def\X{\mathbf{X}}
\def\H{\mathbf{H}}
\def\W{\mathbf{W}}

\def\Dn{\mathbf{D}^{-\frac{1}{2}}}

\def\nbeta{{1-\beta}}

\def\z{\bm{z}}

\def\h{\bm{h}}
\def\dl{\bm{\delta}}

\def\b{\bm{b}}
\def\hsf{\hspace{-0.5mm}}

\def\vvarphi{\bm \varphi}

\def\revision{}
\def\wait{}

\newcommand{\eat}[1]{}

\begin{document}
\title{Decoupled Graph Neural Networks for Large Dynamic Graphs}
\subtitle{[technical report]}

\author{Yanping Zheng}
\affiliation{%
  \institution{Renmin University of China}
  \city{Beijing}
  \state{China}
}
\email{zhengyanping@ruc.edu.cn}

\author{Zhewei Wei}
\authornote{Zhewei Wei is the corresponding author. The work was partially done at Gaoling School of Artificial Intelligence, Peng Cheng Laboratory, Beijing Key Laboratory of Big Data Management and Analysis Methods and MOE Key Lab of Data Engineering and Knowledge Engineering.}
\affiliation{%
  \institution{Renmin University of China}
  \city{Beijing}
  \country{China}
}
\email{zhewei@ruc.edu.cn}

\author{Jiajun Liu}
\affiliation{%
  \institution{Data 61, CSIRO}
  \city{Pullenvale, Queensland}
  \country{Australia}
}
\email{jiajun.liu@csiro.au}

\begin{abstract}
Real-world graphs, such as social networks, financial transactions, and recommendation systems, often demonstrate dynamic behavior. This phenomenon, known as graph stream, involves the dynamic changes of nodes and the emergence and disappearance of edges. To effectively capture both the structural and temporal aspects of these dynamic graphs, dynamic graph neural networks have been developed. However, existing methods are usually tailored to process either continuous-time or discrete-time dynamic graphs, and cannot be generalized from one to the other. In this paper, we propose a decoupled graph neural network for large dynamic graphs, including a unified dynamic propagation that supports efficient computation for both continuous and discrete dynamic graphs. Since graph structure-related computations are only performed during the propagation process, the prediction process for the downstream task can be trained separately without expensive graph computations, and therefore any sequence model can be plugged-in and used. As a result, our algorithm achieves exceptional scalability and expressiveness. We evaluate our algorithm on seven real-world datasets of both continuous-time and discrete-time dynamic graphs. The experimental results demonstrate that our algorithm achieves state-of-the-art performance in both kinds of dynamic graphs. Most notably, the scalability of our algorithm is well illustrated by its successful application to large graphs with up to over a billion temporal edges and over a hundred million nodes.
\end{abstract}

\maketitle

\pagestyle{\vldbpagestyle}
\begingroup\small\noindent\raggedright\textbf{PVLDB Reference Format:}\\
\vldbauthors. \vldbtitle. PVLDB, \vldbvolume(\vldbissue): \vldbpages, \vldbyear.\\
\href{https://doi.org/\vldbdoi}{doi:\vldbdoi}
\endgroup
\begingroup
\renewcommand\thefootnote{}\footnote{\noindent
This work is licensed under the Creative Commons BY-NC-ND 4.0 International License. Visit \url{https://creativecommons.org/licenses/by-nc-nd/4.0/} to view a copy of this license. For any use beyond those covered by this license, obtain permission by emailing \href{mailto:info@vldb.org}{info@vldb.org}. Copyright is held by the owner/author(s). Publication rights licensed to the VLDB Endowment. \\
\raggedright Proceedings of the VLDB Endowment, Vol. \vldbvolume, No. \vldbissue\ %
ISSN 2150-8097. \\
\href{https://doi.org/\vldbdoi}{doi:\vldbdoi} \\
}\addtocounter{footnote}{-1}\endgroup

\ifdefempty{\vldbavailabilityurl}{}{
\vspace{.3cm}
\begingroup\small\noindent\raggedright\textbf{PVLDB Artifact Availability:}\\
The source code, data, and/or other artifacts have been made available at \url{\vldbavailabilityurl}.
\endgroup
}

\section{introduction}
\label{sec:intro}
There are several complex networks in the real world, including social networks, transportation networks, biological networks, etc. In these networks, interactions between nodes include a great deal of valuable information, and graphs are regarded as good information carriers for these complicated networks. As a result, a number of graph analysis problems arise, such as link prediction and anomaly identification. Due to their exceptional performance, Graph Neural Networks (GNNs) are recognized as effective tools for resolving these problems. However, most GNNs are designed for static graphs, while networks are constantly evolving over time. Focusing only on static graph information can result in missing crucial details, such as the patterns of network evolution. For example, social networks are characterized by continuous membership changes and shifts in following and unfollowing among users. By analyzing the temporal patterns of dynamic graphs, we can provide recommendations on potential friendships.

In recent years, various works have been developed to address the challenges of modeling and analyzing dynamic graphs. However, these algorithms are often designed for specific types of data. For example, it is challenging to adapt TGAT~\cite{xu2020tgat}, which was established for Continuous-Time Dynamic Graphs (CTDGs), to Discrete-Time Dynamic Graphs (DTDGs)~\cite{liuxue2022encoder}. Similarly, methods developed for DTDGs, such as DySAT~\cite{sankar2020dysat} and STGCN~\cite{yan2018stgcn}, cannot be directly applied to CTDGs. Although we could transform a CTDG into a sequence of snapshots taken at extremely short intervals and then use DTDG methods, the computational expense would be prohibitive. 

\noindent \textbf{Motivation.} GNNs are important algorithms for solving graph-structured problems. The typical GNN layer consists of two modules, feature propagation and prediction, where feature propagation is the primary element impacting performance~\cite{wu2019sgc}. This has led to the development of decoupled GNNs. Several works, such as APPNP~\cite{klicpera2018ppnp} and GBP~\cite{chen2020gbp}, separate the feature propagation and nonlinear transformation operations, achieving significant improvements and scalability. By pre-calculating feature propagation, the need for {\wait complex computation} during model training can be eliminated, saving time and effort. In addition, effective feature propagation can further improve the performance of GNNs. However, most of the existing decoupled models are tailored for learning static graphs, and their adaptation to dynamic graphs is challenging.

\noindent \textbf{Contribution.} Inspired by the decoupled static GNN, we propose a decoupled dynamic GNN method. The propagation progress takes dynamic graphs as input and generates temporal representations of all nodes in the graph. The model then trains for downstream tasks using the results of dynamic propagation, which no longer requires complicated graph computing at this stage and enables the utilization of arbitrary neural network models. Therefore, the computation for graph-structured data in dynamic graphs exists only in the propagation process, allowing for the construction of generic propagation methods for both continuous-time and discrete-time dynamic graphs.

{\revision We observe that CTDG methods, such as TGN~\cite{rossi2020tgn}, keep track of nodes affected by each graph event and adjust their embeddings, avoiding relearning the embeddings of all nodes and conserving computing resources. Since each snapshot is treated as a static graph in DTDG methods, edge deletion and the simultaneous occurrence of multiple graph events are naturally handled.} {\wait Our objective is to develop a novel dynamic graph neural network that combines the strengths of both CTDG and DTDG methods. To achieve this, we introduce incremental node embedding update strategies specifically designed for handling batch graph events. This allows our model to process batch events similar to DTDG methods, while also keeping track of embedding changes akin to CTDG methods. Notably, our update strategy is not limited to adding new edges but also works seamlessly for removing edges.} The main contributions can be summarized as follows:
\begin{itemize}[leftmargin = *]
  \item We propose a decoupled graph neural network for large dynamic graphs, which decouples the temporal propagation and prediction processes on dynamic graphs, enabling us to achieve great scalability and generate {\wait effective} representation. 
  \item {\wait We support} the processing of continuous-time and discrete-time dynamic graphs by designing the generalized dynamic feature propagation. On the other hand, the model can fit various high-pass or low-pass graph filters to obtain a comprehensive temporal representation, by configuring various propagation formulas.
  \item Extensive experiments on seven benchmark datasets demonstrate the effectiveness of our method. Experimental results show that our model outperforms existing state-of-the-art methods. In addition, we evaluate our method on two large-scale graphs to show its excellent scalability.
\end{itemize}

\section{Notations and Preliminary}
\label{sec:pre}
In this section, we first introduce the necessary notations. Then {\wait we provide a concise overview of the classification of dynamic graphs and the common learning tasks associated with them.}

\noindent \textbf{Notations.} A static graph is denoted as $G=(V, E)$, where $V$ is the set of $n$ nodes, and $E$ represents the set of $m$ edges. 
{\revision Let $\A \hspace{-0.5mm}\in\hspace{-0.5mm} \mathbb{R}^{n \times n}$ represent the adjacency matrix of $G$, with entry $\A(i, j)=w_{(i, j)}>0$ being the weight of the edge between node $i$ and $j$, and $\A(i, j)=w_{(i, j)} \hspace{-0.5mm}=\hspace{-0.5mm} 0$ indicates non-adjacency. The degree matrix $\D \hspace{-0.5mm}\in\hspace{-0.5mm} \mathbb{R}^{n \times n}$ is a diagonal matrix defined by $\D(i,i) \hspace{-0.5mm}=\hspace{-0.5mm} d(i) \hspace{-0.5mm}=\hspace{-0.5mm} \sum_{j \in V} w_{(i,j)}$. Each node $i \hspace{-0.5mm}\in\hspace{-0.5mm} V$ has a $d$-dimensional features vector $\x_i$, and all feature vectors form the feature matrix $\X \in \mathbb{R}^{n \times d}$.} \eat{Table~\ref{tab:notations} in Appendix summarises notations used in this paper.}

\begin{figure}[t]
\setlength{\abovecaptionskip}{2mm}
	\begin{small}
		\centering
		\begin{tabular}{cc}	
		\includegraphics[height=36mm]{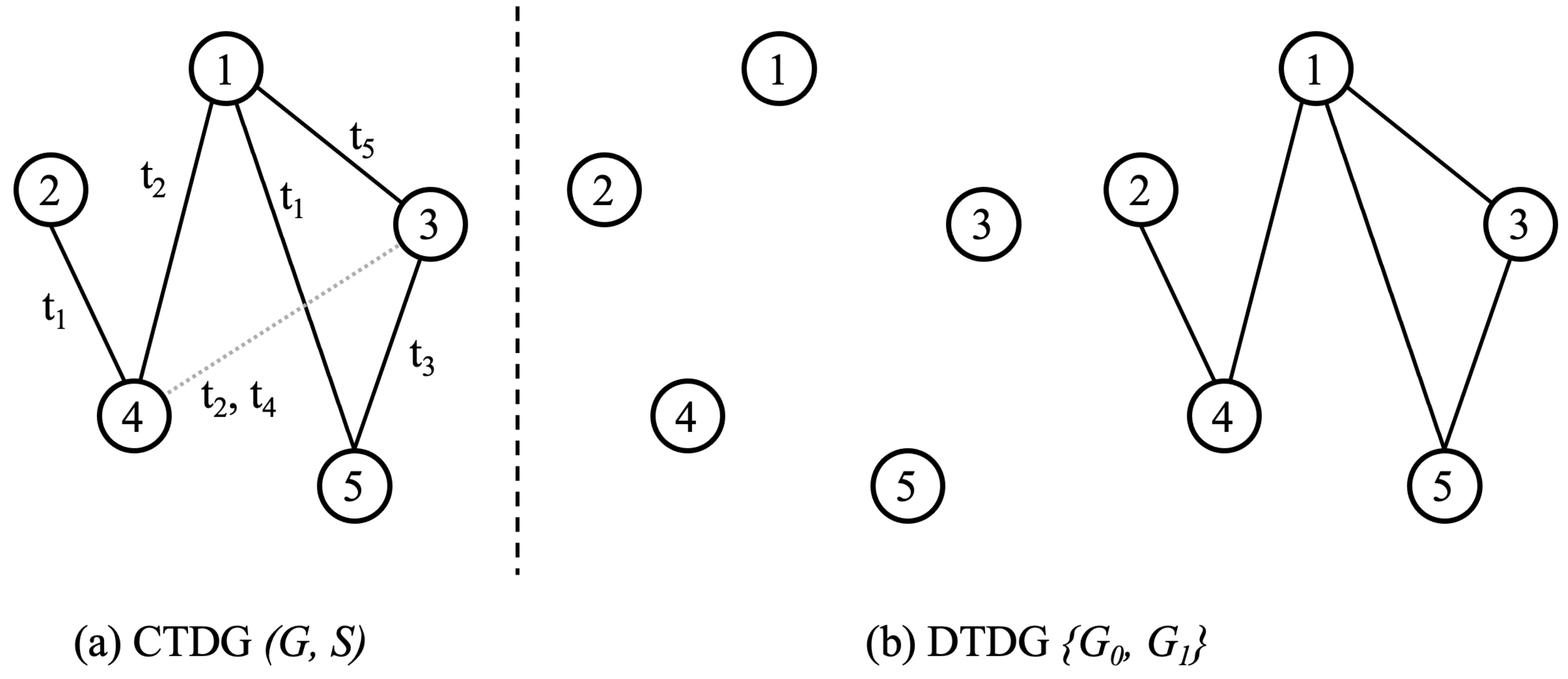} 
		\end{tabular}
		\caption{Two types of dynamic graphs.}
		\label{fig:dynamic_graphs}
	\end{small}
\end{figure}
\noindent \textbf{Dynamic Graphs.} 
Dynamic Graphs can be summarized into two categories, CTDGs and DTDGs, depending on whether the entire timestamp is saved~\cite{Kazemi2020survey}. A CTDG is composed of an initial graph and a sequence of events, denoted as $(G, S)$, where $G$ is the initial state of the dynamic graph at time $t_0$ and $S$ is a set of observed events on the graph. Each event consists of a triplet of {\em (event type, event, timestamp)}, where the {\em event type} can be edge additions, edge deletions, node additions, node deletions, node feature modifications, and so on. Therefore, $G_t$ is the new graph generated from the initial graph $G$ by sequentially completing the graph events of $\{t_1\sim t\}$. Figure~\ref{fig:dynamic_graphs}(a) shows an example of updating from an empty graph with only five nodes to the graph $G_5$ at time $t_5$, where the graph events involved are: 
\begin{equation*}
\begin{aligned}
S =& \left\{{(AddEdge, (v_1, v_5), t_1), (AddEdge, (v_2, v_4), t_1),} \right. \\
&\left.{(AddEdge, (v_1, v_4), t_2), (AddEdge, (v_3, v_4), t_2),} \right.\\
&\left.{(AddEdge, (v_3, v_5), t_3),(DeleteEdge, (v_3, v_4), t_4),} \right.\\
&\left.{(AddEdge, (v_1, v_3), t_5)} \right\}. \\
\end{aligned}
\end{equation*}

{\wait A DTDG is represented as a sequence of snapshots, $\{G_0, \dots, G_T\}$, which are sampled at regular time intervals. Figure ~\ref{fig:dynamic_graphs}(b) illustrates that the second snapshot, $G_1$, of the DTDG can be considered as the graph snapshot captured by the CTDG in Figure~\ref{fig:dynamic_graphs}(a) at time $t_5$. However, it is important to note that the events occurring between $t_1$ and $t_5$ and their respective order are disregarded. Consequently, the DTDG fails to recognize the existence of the previous edge $(v_3, v_4)$ in the graph.}
\eat{
When we state that a dynamic graph is undirected or directed, we mean that at any time $t$, the graph $G_t$ is undirected or directed. A Multi-graph is a graph that contains edges that appear more than once. For example, in a product sales network, a user may buy a particular product many times.}

\noindent \textbf{Graph Learning Tasks.} 
Node classification and link prediction are traditional learning tasks for static graphs. 
{\revision We assume that each node is tagged with a label $\Y(i)$ from the label matrix $\Y$, but only the labels on a subset $V' \subset V$ are known. The objective of the node classification problem is to infer the unknown labels on $V \setminus V'$. }In community detection, for instance, the label assigned to each node represents the community to which it belongs. Link prediction is the classical task of graph learning. It predicts whether an edge exists between two nodes that were not initially connected, {\revision inferring missing edges in $E$. In social networks, link prediction is also known as the friend recommendation task, predicting whether a user is interested in another.}

Similarly, there are node-level and edge-level prediction tasks for dynamic graphs.\eat{ Every time the graph is changed, we update the node representation to reflect the impact of the graph change on the nodes.} Based on the historical information observed so far, we are able to accomplish dynamic node classification and future link prediction defined as follows. 
{\revision
\begin{definition}[Dynamic Node Classification]
\vspace{-2mm}
\label{def:dynamic_node_cls}
For a given graph $G_t=(V_t, E_t)$ and the incomplete label matrix $\Y^\prime_t$, where $G_t$ can be regarded as the graph at timestamp $t$ in a CTDG or the $t$-th snapshot in a DTDG, and $\Y^\prime_t$ associates a subset $V^\prime_t \subset V_t$ with known class labels at timestamp/snapshot $t$, dynamic node classification is to classify the remaining nodes with unknown labels and estimate the label matrix $\Y_t$.
\end{definition}

\begin{definition}[Future Link Prediction]
\vspace{-2mm}
\label{def:future_link_pred}
For a given timestamp/snapshot $t$ and two nodes $i, j\in V_t$, future link prediction aims to predict whether edge $(i, j)$ will be generated in the next timestamp/snapshot or not, based on observations learned from all nodes and their links before timestamp $t$, i.e. observations of $\{G_0, \dots, G_t\}$.
\end{definition}
}

\vspace{-2mm}
\section{Related Works}
\label{sec:related_work}
The Encoder-Decoder framework is a commonly used model in machine learning, which has been applied to various tasks such as unsupervised auto-encoder~\cite{hinton2006science} and neural network machine translation models~\cite{sutskever2014sequence}. Recently, researchers have demonstrated that the Encoder-Decoder framework can generalize most high-performing dynamic graph learning algorithms~\cite{Kazemi2020survey, liuxue2022encoder}.

\subsection{CTDGs Learning Methods} 
It is important to capture the changes in node embedding caused by every graph event when learning CTDGs. Most methods follow the training strategy that the encoder receives a sequence of graph events as input and reflects their influence in node embeddings. The decoder can therefore be a sequence learning model or a static network such as Multilayer Perceptron (MLP) or Support Vector Machine (SVM).

CTDNE~\cite{nguyen2018ctdne} uses the temporal random walk as an encoder and designs three strategies for selecting the next-hop node in dynamic graphs. The introduction of temporal information reduces the uncertainty of embedding, resulting in better performance. The temporal point process is utilized by DyREP~\cite{trivedi2019dyrep} to capture temporal changes at the node and graph levels. DyREP~\cite{trivedi2019dyrep} builds embeddings of target nodes by aggregating information from neighboring nodes, where neighbors are limited by biasing the hop count selection of the temporal point process. These methods inherit the deficiencies of conventional graph representation learning methods, such as their inability to include node properties.

More prevalent CTDG learning encoders are based on Recurrent Neural Networks (RNNs), where the RNN generates memories from observed events associated with the target node via a memory function. The representative model TGN~\cite{rossi2020tgn} comprises a memory component and an embedding component, where the memory component stores the historical memory of the given node. JODIE~\cite{kumar2018jodie}, DyREP~\cite{trivedi2019dyrep}, and TGAT~\cite{xu2020tgat} can be viewed as variants of TGN~\cite{rossi2020tgn}, and they differ in how they update embeddings and memories. Utilizing an asynchronous mail propagator, APAN~\cite{wang2021apan} enforces that graph events are submitted to the model in timestamped order. 
{\revision Wang et al.~\cite{wang2021caw} builds node representations using Causal Anonymous Walks (CAWs), which anonymize the node information on sampled temporal causal routes and apply attention learning to the sampled motifs. The resulting motifs are fed to RNNs to encode each walk as a representation vector. Subsequently, the representations of multiple walks are aggregated into a single vector using a self-attention process for downstream tasks. 

{\wait Generally, methods specific to CTDGs efficiently learn node embeddings by tracking the impacted nodes for each graph event and updating their embeddings accordingly. However, these methods typically focus on considering the immediate neighbors linked with a graph event, such as the endpoints of an inserted edge, and few consider the impact on second-order neighbors~\cite{cui2022dygcn}. Furthermore, the effect on higher-order neighbors or the overall graph is rarely evaluated, and there is limited discussion regarding edge deletions and simultaneous arrivals of multiple events.}}

\vspace{-3mm}
\subsection{DTDGs Learning Methods} 
In the DTDGs learning process, temporal patterns are measured by the sequential relationships between snapshots. Some works apply Kalman filtering~\cite{kempe2000connectivity, sarkar2007latent} or stacked spatial-temporal graph convolution networks (STGCN)~\cite{yan2018stgcn} to create dynamic graph embeddings, and then use simple MLPs as decoders to perform the prediction task. More commonly, static methods, such as GAE and VGAE~\cite{kipf2016vgae}, are used to generate node embeddings of each snapshot. The embeddings are then sorted by time and treated as sequential data, and a sequential decoder is applied to extract the temporal patterns from them.

To obtain embeddings of each snapshot, GraRep~\cite{cao2015grarep}, HOPE~\cite{ou2016asymmetric}, and M-NMF~\cite{wang2017mnmf} construct encoders using matrix decomposition, while DeepWalk~\cite{perozzi2014deepwalk} and node2vec~\cite{grover2016node2vec} transform the graph structure into node-level embeddings using random walk. These algorithms, however, are shallow embedding methods, meaning that they do not consider the attribute information of the graph. Also, there is no parameter sharing between nodes, which makes these methods computationally inefficient. Graph neural networks are efficient ways for learning both the structure and attribute information of a graph. GNNs follow the message-passing framework, in which each node generates embeddings by aggregating information of neighbors~\cite{gilmer2017neural}. To improve the efficiency, Graph Convolutional Network (GCN)~\cite{kipf2017gcn} derives the layer-by-layer propagation formula from the first-order approximation of the localized spectral filters on the graph:
\begin{equation}
\label{equ:gcn_layer}
\H^{(\ell + 1)} = \sigma \left( \Dn \A \Dn \H^{(\ell)} \W^{(\ell)} \right)
\end{equation}
where $\A$ and $\D$ are the adjacency matrix and degree matrix, respectively. $\W^{(\ell)}$ is the learnable parameter of layer $\ell$, and $\sigma$ is a nonlinear activation function such as ReLU. $\H^{(\ell)}$ is the learnt node representation at the $\ell$-th layer, and $\H^{(0)}=\X$. AddGraph~\cite{zheng2019addgraph} employs GCN as the encoder to analyze the structural information of each snapshot, while a sequence decoder is used to determine the relationships between snapshots. \eat{Similarly, Manessi et al.~\cite{manessi2020dynamic} proposed WD-GCN/CD-GCN combined with a variant of Long Short-Term Memory (LSTM)~\cite{hochreiter1997lstm} and extended graph convolution operations to model graph structures and their long  short-term dependencies. }Graph Attention Network (GAT)~\cite{veličković2018gat} is an attention mechanism based on GCN that assigns various weights to the features of neighbors via weighted summation. DyGAT~\cite{sankar2020dysat} employs GAT as an encoder for DTDGs learning, and node embeddings are generated by jointly computing self-attentions of neighborhood structure and time dimensions.

Long Short-Term Memory (LSTM)~\cite{hochreiter1997lstm} is a widely used sequence model, known for its ability to effectively capture long-term temporal dependencies and correlations. Therefore, Seo et al.~\cite{seo2018structured} and Manessi et al.~\cite{manessi2020dynamic} use LSTM as decoders, and their encoders are GCNs or their different versions. The combination structure of GNNs and LSTMs has demonstrated its efficacy in object detection~\cite{yuan2017detection} and pandemic forecasting~\cite{panagopoulos2021mpnn} areas. EvolveGCN~\cite{pareja2020evolvegcn} uses LSTM and Gate Recurrent Unit (GRU) to update the GCN's parameters at each snapshot since it focuses on the evolution of the GCN's parameters rather than the node representation at each snapshot.
{\revision These DTDG-specific methods typically treat each snapshot as a static graph, making it easy for them to address edge deletion and many simultaneous edges.\eat{ Learning on large-scale dynamic graphs can be accomplished by applying scalable static models, such as GraphSAGE~\cite{hamilton2017graphsage} and SGC~\cite{wu2019sgc}, to each snapshot directly.} However, they are unable to track the particular impact of each graph event on node embedding, and the recomputation of each snapshot is computationally expensive.}

\section{Decoupled Graph Neural Network}
\label{sec:method}
{\revision
\textbf{Overview.} As previously indicated, we aim to design a decoupled GNN with high scalability for dynamic graphs. In addition, we also require the model to operate multi-event arrivals simultaneously and support edge deletion while keeping tracking changes in node embedding, which incorporates the benefits of the CTDGs-specific and DTDGs-specific models. Therefore, inspired by the scalable static GNN framework~\cite{wu2019sgc, bojchevski2020pprgo}, we develop a decoupled GNN for large dynamic graphs, in which the dynamic propagation of the graph is decoupled from the prediction process. 
{\wait To enable efficient computation on large-scale dynamic graphs, we employ dynamic propagation with strict error guarantees (as described in Section~\ref{sec:dynamic_propagation}). This approach eliminates learning parameters in the propagation process, facilitating independent graph propagation for generating temporal representations of all nodes. The prediction process focuses on learning the underlying graph dynamics from the representations of nodes, which does not contain expensive graph computations, enabling the use of arbitrary learning models, as described in Section~\ref{sec:prediction_models}}.

\noindent \textbf{Scalable GNNs.} In order to improve the scalability of GNN models, a line of research tries to decouple the propagation and prediction of conventional GNN layers. The idea behind them is to apply MLPs to batches of nodes simply without taking the graph structure into account, which is proposed by SGC~\cite{wu2019sgc} first. For implementation, the representation matrix $\Z$ is generated first following this general formulation propagation: }
\begin{equation}
\vspace{-1mm}
\label{equ:prop_matrix}
    \Z = \sum_{k=0}^\infty \gamma_k (\D^{-a}\A \D^{-b})^k \X \ ,
\end{equation}
where $\X$ denotes the input feature matrix, $a$ and $b$ are convolution coefficient, $\gamma_k (k=0,1,2,\dots)$ is the weight of the $k$-th step convolution. When $a=b=\frac{1}{2}$ and $\gamma_k = 1$, Equation~\ref{equ:prop_matrix} can be considered a GCN with an infinite number of layers, i.e., a stack of infinite layers of Equation~\ref{equ:gcn_layer}. However, the parameters of each layer are discarded for better scalability. {\revision Therefore, MLPs take the representation matrix $\Z$ as input and trains for downstream tasks. {\wait Mini-batch training can be easily accomplished since node representations can be viewed as distinct input samples for the neural network.}} Numerous models, including APPNP~\cite{klicpera2018ppnp}, SGC~\cite{wu2019sgc}, and GBP~\cite{chen2020gbp}, can be regarded as versions of Equation~\ref{equ:prop_matrix} constructed by choosing different values for $a$, $b$, and $\gamma_k$. By varying $\gamma_k$, Equation~\ref{equ:prop_matrix} can approximate any form of graph filter. For instance, Equation~\ref{equ:prop_matrix} corresponds to a low-pass graph filter when all $\gamma_k (k=0,1,2,\dots)$ satisfy $\gamma_k\geq 0$, and Equation~\ref{equ:prop_matrix} relates to a high-pass filter when $\gamma_k$ is of the form $(-\alpha)^k$ with $\alpha \in (0,1)$. For simplicity, we assume that $a=\beta$, $b=1-\beta$, and the sequence of $\gamma_k$ is a geometric progression with a common ratio $\gamma=\frac{\gamma_{_{k+1}}}{\gamma_{_k}}$ and $0<|\gamma|<1$ in this paper. 

{\wait We aim to extend the previous concept to dynamic graphs. Firstly, we derive temporal representations for all nodes in the graph based on dynamic approximate propagation, which can be efficiently precomputed. Next, we batch the structurally enhanced temporal representations of nodes and feed them into the learning model. This decoupling framework, derived from scalable static GNNs, permits the use of any sequence model while preserving high scalability.}
\begin{algorithm}[t]
\setlength{\abovecaptionskip}{-2mm}
\setlength{\belowcaptionskip}{-2mm}
\caption{{\sc GeneralPropagation}}\label{alg:static}
\SetKwInOut{Input}{Input}
\Input{Graph $G$, weight coefficients $\gamma_k$, convolutional coefficients $\beta$, threshold $r_{max}$, initialized $(\epi, \r)$
}
\While{exist $i \in V$ with $|\r(i)|>r_{max} \cdot d(i)^{1-\beta}$}
{
 $\epi(i)\leftarrow \epi(i) + \gamma_0 \cdot \r(i)$\;
 \For{each $j\in N(i)$}
 {
    $\r(j)\leftarrow \r(j)+ \frac{ \gamma \cdot w_{(i,j)} \cdot\r(i)}{d(i)^{1-\beta} d(j)^{\beta}}$ \;
 }
 $\r(i)\leftarrow 0$\;
}
return $(\epi, \r)$\;
\end{algorithm}

\noindent \textbf{Approximate propagation.} The summation in Equation~\ref{equ:prop_matrix} goes to infinity, which makes it computationally infeasible. Following PPRGo~\cite{bojchevski2020pprgo} and AGP~\cite{wang2021agp}, we consider its approximate version. By representing each dimension of the feature matrix as an $n$-dimensional vector $\x$, the feature matrix can be turned into a sequence of $\{\x_0, ..., \x_{d-1}\}$, where the propagation of each vector \eat{$\x_s (0 \hspace{-0.5mm}\leq\hspace{-0.5mm} s \hspace{-0.5mm}<\hspace{-0.5mm} d)$ }is conducted independently. Equation~\ref{equ:prop_matrix} can therefore be expressed in a equivalent vector form: $\vpi=\sum_{k=0}^\infty \gamma_k (\D^{-\beta}\A \D^{\beta - 1})^k \x$. {\revision As illustrated in Algorithm~\ref{alg:static}, we generalize the propagation algorithm~\cite{zheng2022instantgnn} to a weighted version to support weighted graph neural networks and relax the requirement for positive weight coefficients.} We denote the approximate solution as $\epi$, and the cumulative error of all steps is denoted as $\r$. For initialization, we set $\epi=0$ and $\r=\x$. The propagation starts from the node whose residual exceeds the error tolerance $r_{max}$. Then, the node distributes an equal portion of its residual to its neighboring nodes, and the remainder is transformed into its estimate to record the amount of information already propagated by that node. The feature propagation concludes when the residuals of all graph nodes satisfy the error bound. \eat{Therefore, for each node $i\in V$, we have $|\epi(i)-\vpi(i)|\leq r_{max} \cdot d(i)^{1-\beta}$ after Algorithm~\ref{alg:static} ends. }

The neural network model receives the structurally improved feature matrix $\hat{\Z} = (\epi_0, ..., \epi_{d-1})$ as input and is trained to get the final representation of the nodes based on the subsequent task. For instance, the multi-label node classification task typically uses $\Y=softmax(MLP(\hat{\Z}))$. By decoupling propagation and prediction, the model training complexity is independent of the graph topology, which enhances training efficiency and enables the use of sophisticated prediction networks simultaneously.

\vspace{-2mm}
\subsection{Dynamic Propagation}
\label{sec:dynamic_propagation}
We consider a dynamic graph $\mathcal{G}=\{G_0, G_1, ..., G_T\}$, where each $G_t (t \hspace{-0.5mm}\in\hspace{-0.5mm} [0, T])$ is the graph derived from the initial graph in a CTDG after finishing the graph events before timestamp $t$, or the $t$-th snapshot in a DTDG. {\revision That is, $G_t$ refers to the $t$-th observed status of the dynamic graph. We are not concerned with how $G_t$ is obtained, i.e. how the dynamic graph $\mathcal{G}$ is stored. The overall update procedure is summarized in Algorithm~\ref{alg:dynamic}. We obtain the feature propagation matrix for each $G_t$, and $\Z_t$ is derived iteratively from $\Z_{t-1}$, as in lines 9-19. The estimated vector $\epi$ and residual vector $\r$ inherit the propagation results from the previous time step and make the necessary updates based on the current graph structure.}

To construct sequential representations for all nodes in the dynamic graph, it is necessary to comprehend how to quantify the impact that changes to the network have on every node. Therefore, each node should have its individual observation perspective, with a unique comprehension of each graph modification. To improve the computational efficiency, we propose to incrementally compute the node representation when the graph changes. We start with the following theorem on invariant properties. Due to the page limit, we defer the proof to the technical report~\cite{technical_report}.
\begin{theorem}[The Invariant Property]
\vspace{-2mm}
\label{theorem:invariant}
Suppose $\epi(i)$ is the estimate of node $i$, $\r(i)$ is its residual, and $\x(i)$ is its input feature, for each node $i\in V$, we notice that $\epi(i)$ and $\r(i)$ satisfy the invariant property as follow:
\begin{equation}
\label{equ:invariant}
    \epi(i) + \gamma_{_0} \r(i) = \gamma_{_0} \x(i) + \sum_{j\in N(i)} \frac{\gamma \cdot w_{(i, j)} \cdot \epi(j)}{d(i)^{\beta}d(j)^{1-\beta}}.
\end{equation}
\end{theorem}

\begin{algorithm}[t]
\setlength{\abovecaptionskip}{-2mm}
\setlength{\belowcaptionskip}{-2mm}
\caption{{\sc DynamicPropagation}}\label{alg:dynamic}
\SetKwInOut{Input}{Input}
\Input{Dynamic graph $\mathcal{G}$, weight coefficients $\gamma_k$, convolutional coefficients $\beta$, threshold $r_{max}$, feature matrix $\X^{n \times d}$
}

\textbf{parallel} \For{each column $\x\in \X$}
{
  \tcc{Step 1. Generate estimated and residual vector $\epi$, $\r$ for the initial graph $G$ }
  $\epi \leftarrow 0, \r \leftarrow \x $ \;
  $\epi, \r \leftarrow$ {\sc GeneralPropagation($G$, $\gamma_k$, $\beta$, $r_{max}$, $\epi, \r$)}  \;
  \For{each time step $t \in [1,T]$}
  {
     Updating $G$ with graph events at time $t$\;
     Collect the affected nodes at time $t$ as $V_A$ \;
    \tcc{Step 2. Maintain the estimated and residual vector $\epi$, $\r$ in accordance with the invariant property. } 
    \textbf{parallel} \For{each $u\in V_A$}
    {
      $\epi(u) \leftarrow \epi(u) \cdot \frac{d(u)_t^\nbeta}{d(u)_{t-1}^\nbeta}$ \;
      $\r(u) \leftarrow \r(u)+ \epi(u) \cdot \frac{d(u)_{t-1}^\nbeta - d(u)_t^\nbeta}{\gamma_{_0} \cdot d(u)_t^\nbeta}$ \;
    }
    \textbf{parallel} \For{each $u\in V_A$}
    {
      $\Delta \r(u)\hspace{-0.5mm} \leftarrow\hspace{-0.5mm} (\epi(u) \hspace{-0.5mm}+\hspace{-0.5mm} \gamma_{_0} \r(u) \hspace{-0.5mm}-\hspace{-0.5mm} \gamma_{_0} \bm{x}(u)) \hspace{-0.5mm}\cdot\hspace{-0.5mm} \frac{d(u)_{t-1}^\beta - d(u)_t^\beta}{d(u)_t^\beta}$ \;
      \For{each $v \in N_{add,t}(u)$}
      {
        $\Delta \r(u) \leftarrow \Delta \r(u) + \frac{\gamma\epi(v)}{d(u)_t^\beta d(v)_t^\nbeta}$ \;
      }
      \For{each $v \in N_{delete,t}(u)$}
      {
        $\Delta \r(u) \leftarrow \Delta \r(u) - \frac{\gamma\epi(v)}{d(u)_t^\beta d(v)_t^\nbeta}$ \;
      }
      $\Delta \r(u) \leftarrow \Delta \r(u) / \gamma_{_0} $ \;
      $\r(u) \leftarrow \r(u)+\Delta \r(u)$ \;
    }
    \tcc{Step 3. Propagation on the graph at time $t$. }
    $\epi, \r \leftarrow$ {\sc GeneralPropagation($G$, $\gamma_k$, $\beta$, $r_{max}$, $\epi, \r$)}  \;
  }
}
return Embedding matrix $\hat{\Z}^{n\times d}=(\epi_0, \dots, \epi_{d-1})$ \;
\end{algorithm}
\noindent \textbf{Generalized update rules.} Without loss of generality, we assume that an edge $(u,v)$ with weight $w_{(u,v)}$ is inserted to the graph. According to Equation~\ref{equ:invariant}, the set of affected nodes is $V_A=\{u, w | w\in N(u)\}$. For node $u$, the increment caused by the insertion can be quantified as $(\epi(u) + \gamma_{_0} \r(u) - \gamma_{_0} \x(u))\frac{d(u)^\beta - (d(u)+w_{(u,v)})^\beta }{\gamma_{_0} \cdot (d(u)+w_{(u,v)} )^\beta} + \frac{\gamma w_{(u,v)} \epi(v)}{\gamma_{_0}(d(u)+w_{(u,v)})^\beta d(v)^\nbeta}$, since the degree is updated to $d(u)+w_{(u,v)}$ and a new neighbor $v$ appears. According to the meaning of estimate and residual, we add this increment to the residual of node $u$. Similarly, for each node $w \in N(u)$, $\frac{\epi(u)}{d(u)^\nbeta}$ in its equation will be updated to $\frac{\epi(u)}{(d(u)+w_{(u,v)})^\nbeta}$ as a result of the change of node $u$'s degree. To guarantee that the update time complexity of each insertion is $O(1)$, the following updates are performed to prevent alterations to node $u$'s neighbors: 
\begin{itemize}[leftmargin = *]
  \item $\epi(u) = \frac{d(u)^\nbeta}{(d(u)+w_{(u,v)})\nbeta} \cdot \epi(u)$;
  \item $\r(u) = \r(u) + \frac{\epi(u)}{\gamma_{_0}} \cdot (\frac{d(u)^\nbeta}{(d(u)+w_{(u,v)})^\nbeta} - 1)$. 
\end{itemize}
The detailed calculation process of the update and its batched version can also be found in the technical report~\cite{technical_report}. Since none of variables involved in the equation of other nodes have changed, the increment induced by the insertion of the edge $(u,v)$ is zero from the perspective of node $i \hspace{-0.5mm}\in\hspace{-0.5mm} V, i \hspace{-0.5mm}\neq\hspace{-0.5mm} u$ and $i \hspace{-0.5mm}\notin\hspace{-0.5mm} N(u)$. Algorithm~\ref{alg:static} is then used to propagate this increment from node $u$ to its neighbors, informing other nodes of the change in the graph. 

The preceding procedure {\wait can be easily} generalized to the case of deleting the edge $(u,v)$ with weight $w_{(u,v)}$ by simply replacing $(d(u)+w_{(u,v)})$ with $(d(u)-w_{(u,v)})$.
Therefore, it is unnecessary to recalculate the feature propagation when the graph changes, but rather obtain the current propagation matrix incrementally based on the past calculation result. 
{\revision
In addition, we have Theorem~\ref{theorem:error} to guarantee the error of propagation on each $G_t$.
\begin{theorem}[Error Analysis]
\vspace{-2mm}
\label{theorem:error}
Suppose $\epi_t(i)$ is the estimate of node $i$ at time $t$, $\vpi_t(i)$ is its ground-truth estimate at time $t$, $d(i)_t$ is its degree at time $t$, and $r_{max}$ is the error threshold, for each node $i\in V$, we have $|\vpi_t(i) - \epi_t(i)| \leq r_{max} \cdot d(i)_t^\nbeta$ holds for $\forall t \in \{ 0, 1, \dots, T \}$.
\end{theorem}

\noindent \textbf{Handle CTDGs.} We can utilize the aforementioned update strategy to handle incoming graph events accompanied by either inserting or removing edges. For each arriving edge $(u, v)$, we can ensure that Equation~\ref{equ:invariant} holds at all nodes by updating only the estimate and residual at node $u$, and the time complexity of the update is $O(1)$. Therefore, the above update strategy can be well adapted to sequences of frequently arriving graph events in CTDGs.

\noindent \textbf{Handle DTDGs.} For two successive snapshots $G_{t-1}$ and $G_t$ in a DTDG, we regard the changes between the two snapshots as graph events arrived simultaneously, and it is easy to statistically extract $V_A$. We can then compute exactly the increment between the two snapshots by substituting $w_{(u,v)}$ with the degree change $\Delta d(u) = d(u)_t - d(u)_{t-1}$ for each affected node $u \hspace{-0.5mm}\in\hspace{-0.5mm} V_A$. 
}
Similarly, {\wait Algorithm~\ref{alg:static} transmits information about the changes in the graph to other nodes}. Therefore, when a new snapshot $G_t$ arrives, we incrementally update the feature propagation matrix based on the feature propagation results of snapshot $G_{t-1}$. Since batches of graph updates can be efficiently processed, our method naturally supports DTDGs and maintains tracking for underlying node embeddings. 

{\revision
\noindent \textbf{Remark.} {\wait In comparison to CTDG-specific methods, our approach updates the dynamic graph based on timestamps rather than relying solely on the order of each edge. Since it is common for multiple graph events to occur simultaneously at a single time step in real-world scenarios, handling each event individually would be suboptimal. In contrast to DTDG-specific methods, where each graph snapshot is treated as a static graph, we adopt an incremental update approach, which allows us to update the graph incrementally based on the differences between two successive snapshots. Based on this strategy, we avoid recalculating the underlying node embeddings for each snapshot and disregarding changes to them.}
}

\vspace{-2mm}
\subsection{Prediction}
\label{sec:prediction_models}
{\wait In this section, we provide illustrations of the prediction phase by considering dynamic node classification and future link prediction as examples. These two tasks are defined in detail in Section~\ref{sec:pre}.}

\noindent \textbf{Dynamic node classification.} We can incrementally obtain the feature propagation matrix at each {\wait time $t$} using Algorithm~\ref{alg:dynamic}. Each row of the feature propagation matrix $\hat{\Z}_t$, denoted as a $d$-dimensional vector $\z_{t,i}$, is the structural-enhanced node representation vector of node $i\in V$ at time $t$. Therefore, $\z_{t,i}$ is the representation vector of node $i$ under the error tolerance control described in Theorem~\ref{theorem:error}. Since the aggregation of node features based on graph structure has been completed during the propagation process, the node representation vector $\z_{t,i} (t\hsf=\hsf1,\dots, T)$ for each node $i\hsf\in\hsf V$ can be regarded as the standard input vector of neural networks at this stage. For instance, we use a two-layer MLP to predict the label of node $i$ at time $t$ as $\Y_t(i)=softmax(MLP(\z_{t,i}))$.

\noindent \textbf{Future link prediction.} In this task, we aim to learn the temporal pattern of each node to forecast if the {\wait two given nodes} would be linked at the given time. The changes in the node representation over time can be regarded as a time series, and the temporal information contained within it can be captured by a common temporal model such as LSTM. Notice that utilizing sequence $\{\z_{1,i}, \dots, \z_{T,i} \}$ to describe the dynamic network will provide a very subjective impression from node $i$. As a result, when changes in the graph have a large influence on node $i$, the representation vector $\z_{t,i}$ changes significantly with respect to the previous moment $\z_{t-1, i}$. The vector changes little from that of the previous state $\z_{t-1,i}$ when changes in the graph have little influence on node $i$. Note that the degree of influence is related to the final feature propagation matrix generated by Algorithm~\ref{alg:dynamic}. Therefore, node $i$'s perception of the degree of graph change is influenced by the descriptions of its neighboring nodes through the propagation process. {\wait The completion of the future link prediction task involves the following three steps.}
\begin{itemize}[leftmargin = *]
\item \textbf{Firstly}, we calculate the difference between each node's state in two consecutive graph states as $\dl_{t,i}=g(\z_{t,i}, \z_{t-1,i})$, where $g(\cdot)$ is distance measure function. We implement $g(\cdot)$ as a simple first-order distance, although it could also be a $\ell_2$-norm, cosine similarity, or other complicated design. Based on the above, we interpret $\dl_{t,i}(s)$ as the score of graph changes from the perspective of node $i$ at the $s$-th feature dimension. 

\item \textbf{Secondly}, sequence models, such as LSTM, directly take the sequence $\{\dl_{1,i}, \dots, \dl_{t,i}\}$ as input to capture the temporal patterns for node $i$. Since the graph structure information is already included in $\z_{t,i}$ and $\dl_{t,i}$, the sequence model can be employed more effectively by focusing solely on temporal patterns. The predicted state at time $t$ is denoted as $\h_t=\mathcal{M}(\h_{t-1}, \dl_{t})$, where $\mathcal{M}$ is the chosen sequence learning model, $\dl_t$ is the current input vector, and $\h_{t-1}$ is the learned prior state. The standard LSTM cell is defined by the following formula:
\begin{align}
\vspace{-5mm}
    \bm{i}_t &= \sigma(\W_{\bm{i}}\h_{t-1} + \U_{\bm{i}}\dl_{t} +\b_{\bm{i}}), \nonumber \\
    \bm{f}_t &= \sigma(\W_{\bm{f}}\h_{t-1} + \U_{\bm{f}}\dl_{t} +\b_{\bm{f}}), \nonumber \\
    \bm{o}_t &= \sigma(\W_{\bm{o}}\h_{t-1} + \U_{\bm{o}}\dl_{t} +\b_{\bm{o}}), \\
    \bm{\tilde{c}}_t &= \tanh (\W_{\bm{c}}\h_{t-1} + \U_{\bm{c}}\dl_{t} +\b_{\bm{c}}), \nonumber \\
    \bm{c}_t &= \bm{f}_t \odot \mathbf{C}_{t-1} + \bm{i}_t \odot \bm{\tilde{c}}_t, \nonumber \\
    \h_t &= \bm{o}_t \odot \tanh(\bm{c}_t) \nonumber,
\end{align}
where $\sigma$ is the sigmoid activation function, $\odot$ denotes the matrix product operation, $\bm{i}_t$, $\bm{f}_t$ and $\bm{o}_t$ represent the degree parameters of the input gate, forgetting gate and output gate of the LSTM cell at time $t$. $\{\W_{\bm{i}}, \U_{\bm{i}}, \b_{\bm{i}}\}$, $\{\W_{\bm{f}}, \U_{\bm{f}}, \b_{\bm{f}}\}$, $\{\W_{\bm{o}}, \U_{\bm{o}}, \b_{\bm{o}}\}$ are their corresponding network parameters, respectively. $\bm{\tilde{c}}_t$ denotes the candidate states used to update the cell states. $\{\W_{\bm{c}}, \U_{\bm{c}}, \b_{\bm{c}}\}$ are the parameters of the network for generating candidate memories. $\bm{c}_t$ is formed as the output vector $\h_t$ at the current time $t$ after the output gate has discarded some information. Note that the LSTM cell could be replaced by a GRU cell or a Transformer cell, as $\mathcal{M}$ is free from graph-related computations.

\item \textbf{Finally}, we combine the pair of hidden states of node $i$ and $j$ as $\vvarphi_t(i,j) = f(\h_{t,i}, \h_{t,j})$, where $f(\cdot)$ is the combine function, and we experiment on concatenation following previous work~\cite{rossi2020tgn,you2022roland}. Then the probability score of edge $(i, j)$'s existence at time $t$ is given by $\Y_t(i,j)=\sigma(MLP(\vvarphi_t(i,j)))$.
\end{itemize}

\vspace{-2mm}
\section{Experiments}
\label{sec:exp}
In this section, we evaluate the effectiveness of our method on two representative tasks, future link prediction and dynamic node classification, on both CTDGs and DTDGs. Furthermore, we conduct experiments on two large-scale dynamic graphs to demonstrate the scalability of our method.
\begin{table*}[t]
\caption{Statistics of the datasets.}
  \label{tab:dataset}
\vspace{-3mm}
\begin{tabular}{lllllll}
\hline
                 & \#nodes     & \#edges       & max($t$) & \#classes & \#node features & \#edge features \\ \hline
Wikipedia        & 9,227       & 157,474       & 152,757  & 2         & 172 (random)    & 172             \\
Reddit           & 11,000      & 672,447       & 669,065  & 2         & 172 (random)    & 172             \\
UCI-Message      & 1,899       & 59,835        & 87       & -         & 128 (random)    & -               \\
Bitcoin-OTC      & 5,881       & 35,592        & 138      & -         & 128 (random)    & 1               \\
Bitcoin-Alpha    & 3,783       & 24,186        & 138      & -         & 128 (random)    & 1                \\ 
GDELT            & 16,682      & 191,290,882   & 170,522  & 81        & 413             & 186             \\
MAG              & 121,751,665 & 1,297,748,926 & 120      & 152       & 768             & -               \\ \hline
\end{tabular}
\end{table*}

\noindent \textbf{Datasets.} We conducted experiments on seven real-world datasets, including Wikipedia~\cite{kumar2018jodie}, Reddit~\cite{kumar2018jodie}, UCI-MSG~\cite{panzarasa2009uci}, Bitcoin-OTC~\cite{kumar2016bitcoin, kumar2018bitcoin}, Bitcoin-Alpha~\cite{kumar2016bitcoin, kumar2018bitcoin}, GDELT~\cite{zhou2022tgl} and MAG~\cite{zhou2022tgl, hu2021ogblsc}. The statistics of datasets are presented in Table~\ref{tab:dataset}. In all graphs, the weight of an edge is determined by its frequency of occurrence. More details about the datasets can be found in the technical report~\cite{technical_report}.

\noindent \textbf{Baseline methods.} We compare our method to state-of-the-art dynamic graph neural networks, including TGN~\cite{rossi2020tgn}, CAW-Ns~\cite{wang2021caw} for CTDGs and ROLAND~\cite{you2022roland} for DTDGs. In the two CTDG datasets, Wikipedia and Reddit, we strictly inherit the baseline results from their papers and follow the experimental setting of TGN~\cite{rossi2020tgn}. In the three DTDG datasets, UCI-Message, Bitcoin-Alpha, and Bitcoin-OTC, our experimental setting is closely related to those of EvolveGCN~\cite{pareja2020evolvegcn} and ROLAND~\cite{you2022roland}, and we adopt the original paper's stated results. To provide a fair comparison, we employ the same data processing and partitioning techniques as TGN~\cite{rossi2020tgn} and ROLAND~\cite{you2022roland}. For the two large-scale datasets GDELT and MAG, \eat{where previous work could not be trained directly, }we utilize the results reported by TGL~\cite{zhou2022tgl}. Other baseline methods are described in Section~\ref{sec:related_work}.

\subsection{Experiments on CTDGs}
\label{sec:exp_ctdg}
\textbf{Experimental Setting.} We conduct experiments on Wikipedia and Reddit dataset in both transductive and inductive settings, following~\cite{rossi2020tgn}. In both settings, the first 70\% of edges are used as the training set, 15\% are used as the validation set, and the remaining 15\% are used as the test set. In the transductive setting, we predict future links for observed nodes in the training set. In the inductive setting, the future linking status of nodes that do not present in the training set is predicted. {\wait We formulate the prediction of future links between two nodes as a binary classification problem. More specifically, we assign a label of 1 to indicate that the two nodes will be connected in the future, while a label of 0 signifies that there will be no link between them.} The time span of the prediction is one time step. The popular classification metric Average Precision (AP) is employed to evaluate the algorithm's performance on both future link prediction and dynamic node classification tasks. In order to maintain the balance of the data, we generate one negative sample for each test edge or node when computing AP, following the experimental setting in TGN~\cite{rossi2020tgn} and TGL~\cite{zhou2022tgl}. For our method, we set the weight coefficients $\gamma_k = \alpha(1-\alpha)^k$, which is known as the Personalized PageRank weights with a hyperparameter $\alpha \in (0,1)$. {\wait The standard LSTM is utilized as the sequence model to learn the temporal patterns present in the node representation.} Since no node features are provided\eat{ in Wikipedia or Reddit datasets}, we use a randomly generated 172-dimensional vector as the initial node feature vector.

\begin{table}[t]
\caption{Future link prediction on CTDGs. AP (\%) $\pm$ standard deviations computed of 10 random seeds are exhibited.}
  \label{tab:exp_ctdg}
\vspace{-3mm}
\setlength\tabcolsep{2.3pt}
\begin{tabular}{c|cc|cc}
\hline
 & \multicolumn{2}{c}{Wikipedia} & \multicolumn{2}{|c}{Reddit} \\ \hline
              & Transductive         & Inductive            & Transductive & Inductive   \\ \hline
GAE           & 91.44 ± 0.1          & -                    & 93.23 ± 0.3  & -           \\ 
VGAE          & 91.34 ± 0.3          & -                    & 92.92 ± 0.2  & -           \\ 
DeepWalk      & 90.71 ± 0.6          & -                    & 83.10 ± 0.5  & -           \\ 
Node2Vec      & 91.48 ± 0.3          & -                    & 84.58 ± 0.5  & -           \\ 
GAT           & 94.73 ± 0.2          & 91.27 ± 0.4          & 97.33 ± 0.2  & 95.37 ± 1.1 \\ 
GraphSAGE     & 93.56 ± 0.2          & 91.09 ± 0.3          & 97.65 ± 0.2  & 96.27 ± 0.2 \\ 
CTDNE         & 92.17 ± 0.5          & -                    & 91.41 ± 0.3  & -           \\ 
Jodie         & 94.62 ± 0.5          & 93.11 ± 0.4          & 97.11 ± 0.3  & 94.36 ± 1.1 \\ 
TGAT          & 95.34 ± 0.1          & 93.99 ± 0.3          & 98.12 ± 0.2  & 96.62 ± 0.3 \\ 
DyRep         & 94.59 ± 0.2          & 92.05 ± 0.3          & 97.98 ± 0.1  & 95.68 ± 0.2 \\ 
TGN           & 98.46 ± 0.1    & 97.81 ± 0.1    &  98.70 ± 0.1  & 97.55 ± 0.1 \\ 
{\revision CAW-N-mean}         & {\revision 98.82 ± 0.1}    & {\revision 98.28 ± 0.1}    &  {\revision 98.72 ± 0.1}  & {\revision 98.74 ± 0.1} \\
{\revision  CAW-N-attn}        & {\revision \ul 98.84 ± 0.1}    & {\revision \ul 98.31 ± 0.1}    & {\revision \ul 98.80 ± 0.1}  & {\revision \ul 98.77 ± 0.1} \\
\textbf{ours} & \textbf{99.16 ± 0.3} & \textbf{98.54 ± 0.2} & \textbf{99.51 ± 0.5}   & \textbf{98.81 ± 0.6} \\ \hline
\end{tabular}
\vspace{-3mm}
\end{table}
\begin{table}[t]
\caption{Dynamic node classification on CTDGs. ROC AUCs (\%) are exhibited.}
  \label{tab:nodecls}
\begin{tabular}{c|cc}
\hline
              & Wikipedia      & Reddit         \\ \hline
Jodie         & 81.37          & \textbf{70.91} \\ 
DySAT         & 86.30          & 61.70          \\
TGAT          & 85.18          & 60.61          \\
TGN           & {\ul 88.33}    & 63.78          \\
APAN          & 82.54          & 62.00          \\ 
\textbf{ours} & \textbf{89.81} & {\ul 67.53}    \\ \hline
\end{tabular}
\vspace{-4mm}
\end{table}
\begin{table*}[t]
\caption{Future link prediction on DTDGs. MRR $\pm$ standard deviations computed of 3 random seeds are exhibited.}
  \label{tab:exp_dtdg}
 \vspace{-3mm}
\begin{tabular}{cc|ccc}
\hline
                         &                & UCI-Message                                  & Bitcoin-Alpha                                & Bitcoin-OTC                                  \\ \hline
\multicolumn{2}{c|}{GCN}                   & 0.1141                                       & 0.0031                                       & 0.0025                                       \\
\multicolumn{2}{c|}{DynGEM}                & 0.1055                                       & 0.1287                                       & 0.0921                                       \\
\multicolumn{2}{c|}{dyngraph2vecAE}        & 0.0540                                       & 0.1478                                       & 0.0916                                       \\
\multicolumn{2}{c|}{dyngraph2vecAERNN}     & 0.0713                                       & 0.1945                                       & 0.1268                                       \\
\multicolumn{2}{c|}{EvolveGCN-H}           & 0.0899                                       & 0.1104                                       & 0.0690                                       \\
\multicolumn{2}{c|}{EvolveGCN-O}           & 0.1379                                       & 0.1185                                       & 0.0968                                       \\
\multicolumn{2}{c|}{ROLAND Moving Average} & 0.0649 ± 0.0049                              & 0.1399 ± 0.0107                              & 0.0468 ± 0.0022                              \\
\multicolumn{2}{c|}{ROLAND MLP}            & 0.0875 ± 0.0110                              & 0.1561 ± 0.0114                              & 0.0778 ± 0.0024                              \\
\multicolumn{2}{c|}{ROLAND GRU}            & {\ul 0.2289 ± 0.0618 }                             & 0.2885 ± 0.0123                              & 0.2203 ± 0.0167                              \\ \hline
\multirow{3}*{\textbf{ours}}  & GRU            & 0.2024 ± 0.0010          & {\ul 0.3289 ± 0.0070}    & 0.2985 ± 0.0121          \\
  & LSTM           & 0.2140 ± 0.0034    & \textbf{0.3405 ± 0.0133} & {\ul 0.3102 ± 0.0046}    \\
  & Transformer    & \textbf{0.2314 ± 0.0048} & 0.3173 ± 0.0135          & \textbf{0.3110 ± 0.0049} \\\hline
\end{tabular}
\end{table*}
\noindent \textbf{Results.} The results of future link prediction in both transductive and inductive settings are shown in Table~\ref{tab:exp_ctdg}. The presented results are the average of 10 runs. Our method performs better than baseline methods in both transductive and inductive settings. The interesting thing is that we did not use the provided edge features and achieve comparable or even better performance. This may be strongly related to the experimental setting and dataset. For the current future link prediction, we simply need to forecast whether a connection will be created between two given nodes in the future. The specifics of that link are practically of no concern. The publicly accessible edge features of Wikipedia and Reddit are derived from the textual content of each edit or post on the respective web page and sub-reddit. The learning objective is to detect whether a user would edit a certain page or post on a given sub-reddit in the future, without predicting the edit or post content. It is possible that semantic information of textual material is superfluous. The historical interaction data already contains sufficient information to reveal users' preferences for particular pages and sub-reddits. Our hypothesis is also supported by the results of our method on graphs that lack semantic information.

Table~\ref{tab:nodecls} shows the experimental results for the dynamic node classification. For node classification, we always use the most recent node representation based on the history observed so far. The three-layer MLP is employed as the classifier. The results in Table~\ref{tab:nodecls} show that our method effectively captures the temporal changes of the nodes in time, thus enabling the correct classification of the nodes.


\vspace{-3mm}
\subsection{Experiments on DTDGs}
\label{sec:exp_dtdg}
\textbf{Experimental Setting.} {\wait We use three datasets in this experiment: Bitcoin-OTC, Bitcoin-Alpha and UCI-Message. To ensure a fair comparison, we partition the dataset and calculate evaluation measures in the same manner as ROLAND~\cite{you2022roland}. Since node features and edge features are not provided in these three datasets, we generate the 128-dimensional random vector to serve as the initial node feature. The ranking metric, Mean Reciprocal Rank (MRR), is employed to evaluate performance. We collect 1000 negative samples for each positive sample and then record the ranking of positive samples according to predicted probabilities. MRR is calculated independently for each snapshot in the test set, and the average of all snapshots is reported. For our method, we combine the node temporal representations obtained under settings $\gamma_k\hspace{-0.5mm}=\alpha(1\hspace{-0.5mm}-\hspace{-0.5mm}\alpha)^k$ and $\gamma_k\hspace{-0.5mm}=\alpha(\alpha\hspace{-0.5mm}-\hspace{-0.5mm}1)^k$ to approximate the low-pass and high-pass filters on the graph and introduce low-frequency and high-frequency information, respectively. Three traditional sequence models, LSTM~\cite{hochreiter1997lstm}, GRU~\cite{cho2014gru} and Transformer~\cite{vaswani2017transformer}, are used to finish the future link prediction task.}

\noindent \textbf{Results.} The results in Table~\ref{tab:exp_dtdg} demonstrate the state-of-the-art performances of our method. In the Bitcoin-OTC dataset, our method outperforms the second-best method ROLAND by 41\%. The results show that the LSTM model consistently outperforms the GRU model, which is possibly due to the LSTM having more parameters. The Transformer model tends to achieve a higher MRR, which could be attributed to its holistic approach to the temporal sequence of nodes and reduced reliance on previous hidden states. Additionally, we conducted an ablation study in the technical report~\cite{technical_report} to validate the necessity of introducing high-frequency information.

\begin{table}[t]
\caption{Dynamic node classification on large graphs. F1-Micros (\%) are exhibited.}
  \label{tab:large_nodecls}
\vspace{-3mm}
\begin{tabular}{c|cc}
\hline
      & GDELT          & MAG            \\ \hline
Jodie & 11.25          & 43.94          \\
DySAT & 10.05          & 50.42          \\
TGAT  & 10.04          & {\ul 51.72}    \\
TGN   & {\ul 11.89}    & 49.20          \\
APAN  & 10.03          & -              \\ 
\textbf{ours}  & \textbf{25.49} & \textbf{61.40} \\ \hline
\end{tabular}
\vspace{-3mm}
\end{table}
\vspace{-2mm}
\subsection{Experiments on Large Graphs}
\textbf{Experimental Setting.} To demonstrate the scalability of our method, we conduct experiments on two large-scale real-world graphs, GDELT and MAG. {\revision We exclude EvolveGCN~\cite{pareja2020evolvegcn}, ROLAND~\cite{you2022roland} and CAW-Ns~\cite{wang2021caw} from experiments on GDELT and MAG datasets, since they met out-of-memory issues on both datasets.} Note that the scalability of the baseline methods JODIE~\cite{kumar2018jodie}, DySAT~\cite{sankar2020dysat}, TGAT~\cite{xu2020tgat}, TGN~\cite{rossi2020tgn}, and APAN~\cite{wang2021apan} was not taken into account in their original papers, and their original versions cannot be trained on these two large-scale dynamic graphs. TGL~\cite{zhou2022tgl} has successfully applied these methods to large-scale graphs by developing a distributed dynamic graph neural network training framework. In contrast, our method can learn large-scale graphs directly. 
Our method exhibits greater scalability due to the elimination of parameters in the propagation process, allowing the training on these two large-scale graphs to be completed on a single machine.
We validate the performance of all methods on the dynamic node classification task and compare their performance using the multiple-class classification metric F1-Micro. To guarantee a fair comparison, we ensure that the training, validation and test sets are consistent with the settings in TGL. For our method, we set $\gamma_k = \alpha (1-\alpha)^k$ to obtain the temporal representation of each node, and a three-layer MLP is utilized to complete the training for the classification task.

\noindent \textbf{Results.} Table~\ref{tab:large_nodecls} shows the dynamic node classification results. Compared to baseline methods, we achieve significant performance improvement in both datasets. Specifically, our method improves F1-Micro by 13.6 on the GDELT dataset and 9.68 on the MAG dataset. This indicates that our method can effectively capture the dynamic changes in node representations by precisely locating the directly affected nodes via Equation~\ref{equ:invariant} and quantifying the degree of graph change. The following propagation process that immediately follows broadcasts the change from the affected node to its surroundings, so that higher-order neighbors can also naturally perceive the change on the graph. 
However, from a practical application standpoint, the performance of all methods in GDELT is not adequate. We note that this is due to the presence of much noise in the labeled data of GDELT. Participants can participate in events held around the world via remote means such as online, resulting in some nodes may simultaneously belong to many classes.

\vspace{-2mm}
\section{conclusion}
This paper propose a universal general graph neural network for dynamic graphs that can extract the structural and attribute information of the graph, as well as the temporal information. Our algorithm is based on the framework of decoupled GNNs, which can pre-compute temporal propagation in dynamic graphs and then train them for downstream tasks depending on the nodes' temporal representation. We devised a unified dynamic propagation methods to support the learning on both continuous-time and discrete-time dynamic graphs. Empirical studies on continuous-time and discrete-time dynamic graphs at various scales demonstrate the scalability and state-of-the-art performance of our algorithm.

\begin{acks}
 This research was supported in part by National Key R\&D Program of China (2022ZD0114802), by National Natural Science Foundation of China (No. U2241212, No. 61972401, No. 61932001, No. 61832017), by the major key project of PCL (PCL2021A12), by Beijing Natural Science Foundation (No. 4222028), by Beijing Outstanding Young Scientist Program No.BJJWZYJH012019100020098, by Alibaba Group through Alibaba Innovative Research Program, and by Huawei-Renmin University joint program on Information Retrieval. Jiajun Liu was supported in part by CSIRO's Science Leader project R-91559. We also wish to acknowledge the support provided by Engineering Research Center of Next-Generation Intelligent Search and Recommendation, Ministry of Education. Additionally, we acknowledge the support from Intelligent Social Governance Interdisciplinary Platform, Major Innovation \& Planning Interdisciplinary Platform for the “Double-First Class” Initiative, Public Policy and Decision-making Research Lab, Public Computing Cloud, Renmin University of China.
\end{acks}

\begin{small}
\bibliographystyle{plain}
\bibliography{paper}
\end{small}

\appendix
\section{Notations}
\begin{table}[h]
\caption{\revision Summary of notations.}
  \label{tab:notations}
\vspace{-3mm}
\begin{tabular}{c|l}
\hline
\textbf{Notation}   & \textbf{Description}        \\ \hline
$G=(V, E)$          & the graph with node set $V$ and edge set $E$    \\
$\A, \D$            & the adjacent and degree matrix         \\
$w_{(i, j)}$        & the weight of edge $(i, j) \in E$                          \\
$N(i)$              & the neighbor set of node $i$                               \\
$d(i)$              & the degree of node $i$, $d(i) = \sum_{j \in N(i)} w_{(i, j)}$ \\
$\x$                & the feature vector                            \\
$\vpi, \epi$        & the true and estimated propagation vectors    \\
$\r$                & the estimated residual vector                 \\
$(G, S)$            & the CTDG with event set $S$                                \\
$\{G_0, ..., G_T\}$ & the DTDG with $T+1$ snapshots                                \\
$G_t$               & the graph at time/snapshot $t$                             \\
$\A_t, \D_t$        & the adjacent and degree matrix of graph $G_t$  \\
$\Y$ $(\Y_t)$       & the node label matrix of graph $G$ ($G_t$) \\\hline
\end{tabular}
\end{table}

\section{Proofs}
\subsection{Proof of Theorem~\ref{theorem:invariant}}
We first rewrite Equation~\ref{equ:invariant} into the vector form:
\begin{equation}
\label{equ:invariant-vector}
    \epi + \gamma_{_0} \r = \gamma_{_0} \x + \gamma \cdot \P \epi ,
\end{equation}
where $\epi$ and $\r$ are the estimated and residual vectors, respectively. $\P = \D^{-\beta} \A \D^{\beta - 1}$ is the propagation matrix, and $\gamma_{_0}$ is is the weight coefficient of the $0$-th step propagation, $\gamma$ is the common ratio of geometric progression $\{\gamma_{_0}, \gamma_{_1}, \gamma_{_2}, \dots \}$.  $\x$ denotes the input feature vector. 
\begin{proof}
For initialization, we set $\epi = \bm{0}$ and $\r= \x$, so Equation~\ref{equ:invariant-vector} holds at the beginning of propagation process. Suppose that after $\ell-1$ steps of propagation, we have $\epi^{(\ell-1)} + \gamma_{_0} \r^{(\ell-1)} = \gamma_{_0} \x + \gamma \cdot \P \cdot \epi^{(\ell-1)} $. At the $\ell$-th step of propagation, there is a node $i$ with residual $\r(i)$ that exceeds the permissible error range. Therefore, according to Algorithm~\ref{alg:static}, $\epi$ and $\r$ will update as follows: 
\begin{align}
\label{equ:pi-r-update}
    \r^{(\ell)} &= \r^{(\ell-1)} - \r^{(\ell-1)}(i) + \gamma \cdot \P \cdot \r^{(\ell-1)}(i) \nonumber \\
    \epi^{(\ell)} &= \epi^{(\ell-1)} + \gamma_{_0} \cdot \r^{(\ell-1)}(i).
\end{align}
In the following, we prove Equation~\ref{equ:invariant-vector} holds after the update. By shifting the terms, $\epi^{(\ell)}$ can be expressed as:
\begin{align}
\label{equ:pi-proof}
    \epi^{(\ell)} =& \gamma_{_0} \x + \gamma \cdot \P \cdot \epi^{(\ell)} - \gamma_{_0} \r^{(\ell)} \nonumber \\
    =& \gamma_{_0} \x + \gamma \cdot \P \cdot \left( \epi^{(\ell-1)} + \gamma_{_0}\r^{(\ell-1)(i)} \right) \nonumber \\ 
    & - \gamma_{_0} \left( \r^{(\ell-1)} - \r^{(\ell-1)}(i) + \gamma \cdot \P \cdot \r^{(\ell-1)}(i) \right) \\
    =& \gamma_{_0} \x + \gamma \cdot \P \cdot \epi^{(\ell-1)} - \gamma_{_0} \r^{(\ell-1)} + \gamma_{_0} \r^{(\ell-1)}(i) \nonumber \\
    =& \epi^{(\ell-1)} + \gamma_{_0} \r^{(\ell-1)}(i) \nonumber \\
    \nonumber
\end{align}
In the last equality, we use the fact: $\epi^{(\ell-1)} = \gamma_{_0} \x + \gamma \cdot \P \cdot \epi^{(\ell-1)} - \gamma_{_0} \r^{(\ell-1)}$. According Equation~\ref{equ:pi-r-update}, we have $\epi^{(\ell-1)} + \gamma_{_0} \r^{(\ell-1)}(i) = \epi^{(\ell)}$, and Equation~\ref{equ:pi-proof} also holds after the update at the $\ell$-th step of propagation. Therefore, we have Equation~\ref{equ:invariant-vector} holds at every $\ell (\ell = 0, 1, 2, \dots)$ step, and Theorem~\ref{theorem:invariant} holds.
\end{proof}

\subsection{Proof of Theorem~\ref{theorem:error} (Error Analysis)}
\label{sec:error_analysis}
Recall that in Algorithm~\ref{alg:dynamic}, we always use the {\sc GeneralPropagation} function of Algorithm~\ref{alg:static} to complete the propagation process after we modified the increment to the affected node. As a result, the error control of the propagation result is entirely handled by the {\sc GeneralPropagation} function.

\begin{proof}
To prove the error bound in Theorem~\ref{theorem:error}, we observe that the GeneralPropagation function terminates only if the residuals of all nodes $i\in V$ in the graph satisfy $|\r_t(i)| < r_{max} \cdot d_t(i)^\nbeta$. According to the definition and Equation~\ref{equ:invariant-vector}, we have:
\begin{equation}
    \left \{
    \begin{aligned}
        &\vpi_t = \gamma \P_t \vpi_t + \gamma_{_0} \x, \\
        &\epi_t = \gamma \P_t \epi_t + \gamma_{_0} (\x - \r_t),
    \end{aligned}
    \right.
\end{equation}
where the above two equations are valid for $\forall t \in \{ 0, 1, \dots, T \}$. By subtracting these two equations, we have:
\begin{equation}
    \begin{aligned}
        & \vpi_t - \epi_t = \gamma \P_t \cdot (\vpi_t - \epi_t) + \gamma_{_0} \r_t \\
        \Leftrightarrow& (\I - \gamma \P_t) \cdot (\vpi_t - \epi_t) = \gamma_{_0} \r_t \\
        \Leftrightarrow& \vpi_t - \epi_t = \gamma_{_0} (\I - \gamma \P_t)^{-1} \cdot \r_t .
    \end{aligned}
\end{equation}
In the last equality, we utilize the assumption that $0<|\gamma|<1$. Therefore, the matrix $(\I - \gamma \P_t)$ is invertible. For only illustration purposes, we denote $\Q_t=\gamma_{_0} (\I - \gamma \P_t)^{-1} = \gamma_{_0} \cdot \sum_{\ell=0}^\infty \gamma^\ell \D_t^\nbeta \cdot (\D_t^{-1}\A_t)^\ell \cdot \D_t^{\beta-1}$, such that the $(i,j)$-th element of $\Q_t$ is $\Q_t(i,j)$. For each node $i$, we have: 
\begin{equation}
    \begin{aligned}
        & \left|\vpi_t(i) - \epi_t(i)\right| \\
        =& \left| \sum_{j \in V} \Q_t(i, j) \cdot \r_t(j) \right| \\
        \leq& \sum_{j \in V} d_t(i)^\nbeta \cdot \gamma_{_0} \sum_{\ell=0}^\infty \gamma^\ell \left(\D_t^{-1} \A_t\right)^\ell \hspace{-1.5mm}(i,j) \cdot d_t(j)^{\beta-1} \cdot r_{max}\cdot d_t(j)^\nbeta \\
        =& r_{max} \cdot d_t(i)^\nbeta \cdot \gamma_{_0} \sum_{\ell=0}^\infty \gamma^\ell \sum_{j \in V} \left(\D_t^{-1} \A_t\right)^\ell (i,j) \\
        =& r_{max} \cdot d_t(i)^\nbeta \cdot \gamma_{_0} \sum_{\ell=0}^\infty \gamma^\ell \\
        =& r_{max} \cdot d_t(i)^\nbeta \cdot \gamma_{_0} \cdot \frac{1}{1-\gamma}.
    \end{aligned}
\end{equation}
In the penultimate equation, we use the fact that $\sum_{j \in V} \left(\D^{-1} \A\right)^\ell (i,j) = 1$. By setting the weight coefficients $\gamma_k$ as $ \gamma = \frac{\gamma_{k+1}}{\gamma_k}, \gamma_{_0} = 1-\gamma$, we have $|\vpi_t(i) - \epi_t(i)|\leq r_{max} \cdot d_t(i)^\nbeta$ and this theorem follows.
\end{proof}

\section{Increments Calculation}
Based on the invariant property of Theorem~\ref{theorem:invariant}, we transform the problem of dynamic propagation into the problem of maintaining the equation of $\epi$ and $\r$ on the dynamic graph. We can obtain feasible $\epi$ and $\r$ for the new graph by locally modifying the affected nodes and keeping the equation holds. Moreover, without sacrificing generality, we assume that an edge $(u, v)$ is inserted. Since just two variables, the degree and neighbor set of node $u$, have changed, we conclude that for the nodes in the graph, only the equation at nodes $u$ and $w\in N(u)$ no longer holds. We consider the equation at node $u$ first, and substitute the $i$ with $u$ to get $\epi(u) + \gamma_{_0} \r(u) = \gamma_{_0} \x(u) + \sum_{j\in N(u)} \frac{\gamma w_{(u, j)} \cdot \epi(j)}{d(u)^\beta d(j)^{1-\beta}}$. After the insertion of the edge $(u, v)$, the equation at node $u$ is modified as follows:
\begin{align}
    \epi(u) + \gamma_{_0} \r(u) + \Delta = &\gamma_{_0} \x(u) + \sum_{j\in N(u)} \frac{\gamma w_{(u, j)} \cdot \epi(j)}{(d(u) + w_{(u, v)})^{\beta}d(j)^{1-\beta}} \\
    & + \frac{\gamma w_{(u, v)} \cdot \epi(v)}{(d(u) + w_{(u, v)})^{\beta}d(v)^{1-\beta}} , \nonumber
\end{align}
where the modification happens on the right-hand side of the equation, and we denote the specific increment by $\Delta$, which is placed on the left side of the equation to maintain the validity of the equation. Subtracting the original equation from the modified equation, we have:
\begin{align}
    \Delta =& \frac{\gamma w_{(u, v)} \cdot \epi(v)}{(d(u) + w_{(u, v)})^{\beta}d(v)^\nbeta} + \sum_{j\in N(u)} \frac{\gamma w_{(u, j)} \cdot \epi(j)}{d(j)^\nbeta} \nonumber \\ 
    & \cdot \left( \frac{1}{(d(u)+w_{(u,v)})^\beta} - \frac{1}{d(u)^\beta} \right) \\
    =& \frac{\gamma w_{(u, v)} \cdot \epi(v)}{(d(u) + w_{(u, v)})^{\beta}d(v)^\nbeta} + \left(\epi(u) + \gamma_{_0} \r(u) -\gamma_{_0} \x(u)\right) \nonumber \\
    & \cdot \frac{d(u)^\beta - (d(u)+w_{(u,v)})^\beta}{(d(u)+w_{(u,v)})^\beta} , \nonumber
\end{align}
where we use the invariant property that $\sum_{j\in N(u)} \frac{\gamma w_{(u,j) \cdot \epi(j)}}{d(u)^\beta d(j)^\nbeta} = \epi(u) + \gamma_{_0} \r(u) -\gamma_{_0} \x(u)$ in the last equality. According to the definitions of estimates and residuals, we accumulate the increment to the residual of node $u$, and since the residual is preceded by a factor $\gamma_{_0}$, we finally modify $\r(u)$ as follows: 
\begin{align}
    \r(u) =& \r(u) + \frac{\gamma w_{(u, v)} \cdot \epi(v)}{ \gamma_{_0} (d(u) + w_{(u, v)})^\beta d(v)^\nbeta} \\
    & + \left(\epi(u) + \gamma_{_0} \r(u) -\gamma_{_0} \x(u)\right) \cdot  \frac{d(u)^\beta - (d(u)+w_{(u,v)})^\beta}{ \gamma_{_0} (d(u)+w_{(u,v)})^\beta}. \nonumber
\end{align}
For each node $w\in N(u)$, since its equation involves $\frac{\gamma w_{w,u} \cdot \epi(u)}{d(w)^\beta d(u)\nbeta}$, and when the degree of node $u$ has changed, the equation will be updated as follows:
\begin{align}
    \epi(w) + \gamma_{_0} \r(w) + \Delta =& \gamma_{_0} \x(w) + \hspace{-0.5mm} \sum_{\substack{ j\in N(w)\\ j\neq u}} \hspace{-1mm} \frac{\gamma w_{(w,j)} \cdot \epi(j)}{d(w)^\beta d(j)^\nbeta} \\
    &+ \frac{\gamma w_{(u,j)} \cdot \epi(u)}{d(w)^\beta (d(u)+w_{(u,v)})^\nbeta} \nonumber
\end{align}
where $\Delta$ is the increment and can be easily calculated as $\frac{\gamma w_{(u,j)} \cdot \epi(u)}{d(w)^\beta} \cdot \left( \frac{1}{(d(u)+w_{(u,v)})^\nbeta} - \frac{1}{d(u)^\nbeta} \right)$. Obviously, we could accumulate this increment to the residual of node $w$ according to the aforementioned update strategy, but then each alteration of node $u$ would necessitate an iteration update of its neighbors, which would increase O(d(u)) complexity and is not conducive to parallelization. Therefore, we update the estimate of node $u$ as $\epi(u) = \epi(u) \cdot \frac{(d(u)+w_{(u,v)})^\nbeta}{d(u)^\nbeta}$ to avoid modifying node $w$'s residual and keep its equation hold. Correspondingly, the residual of node $u$ is modified to $\r(u) + \Delta \r(u) = \r(u) + \epi(u) \cdot (\frac{d(u)^\nbeta}{(d(u)+w_{(u,v)})^\nbeta} -1)$ to ensure that the equation at node $u$ is valid. The equivalence of the two update strategies is straightforward to establish. Since $\Delta$ is numerically equivalent to $\Delta \r(u) \cdot \gamma \cdot \P(w,u)$, where $\P(w,u)$ is the transfer probability for node $w$ to $u$, only one additional step of propagation is required to distribute $\Delta \r(u)$ to its neighbors $w \in N(u)$.

\noindent \textbf{Deletion.} The scenario for deleting the edge $(u, v)$ with weight $w_{(u,v)}$ is substantially identical to the scenario described for adding an edge. Since the degree of node $u$ is changed to $d(u)-w_{(u,v)}$, the increment $\Delta$ needs just replace $d(u)+w_{(u,v)}$ with $d(u)-w_{(u,v)}$, and we update node $u$'s residual as $\r(u) = \r(u) - \frac{\gamma w_{(u, v)} \cdot \epi(v)}{ \gamma_{_0} (d(u) - w_{(u, v)})^\beta d(v)^\nbeta} + \left(\epi(u) + \gamma_{_0} \r(u) -\gamma_{_0} \x(u)\right) \cdot  \frac{d(u)^\beta - (d(u)-w_{(u,v)})^\beta}{ \gamma_{_0} (d(u)-w_{(u,v)})^\beta} $. The estimate and residual of node $u$ are similarly updated as described above to avoid looping over its neighbors. At this point, the update ratio of $\epi(u)$ is $\frac{(d(u)-w_{(u,v)})^\nbeta}{d(u)^\nbeta}$.


\noindent \textbf{Batch update.} Node $u$ has multiple added or deleted neighbors when graph events arrive in bulk, it seems unwise to calculate the increment edge by edge. We extend the aforementioned update strategy such that it can calculate the increments generated by all graph events associated with node $u$ concurrently for all node $u \in V_A$, where $V_A$ represents the affected node set. Denoting the set of added neighbors of node $u$ as $N_{add}(u)$ and the set of removed neighbors as $N_{delete}(u)$, we update the estimate and residual of node $u$ as follows: 
\begin{align}
    \epi(u) =& \epi(u) \cdot \frac{(d(u)+\Delta d(u))^\nbeta}{d(u)^\nbeta}, \\
    \r(u) =& \r(u) + \epi(u) \cdot \left(\frac{d(u)^\nbeta}{(d(u)+\Delta d(u))^\nbeta} - 1 \right),  \\
    \r(u) =& \r(u) + (\epi(u) + \gamma_{_0} \r(u) - \gamma_{_0} \x(u))\frac{d(u)^\beta - (d(u)+\Delta d(u))^\beta }{\gamma_{_0} \cdot (d(u)+\Delta d(u))^\beta} \nonumber \\
    & + \sum_{v \in N_{add}(u)} \frac{\gamma w_{(u,v)}\epi(v)}{\gamma_{_0} (d(u)+\Delta d(u))^\beta d(v)^\nbeta} \\
    & - \sum_{w\in N_{delete}(u)} \frac{\gamma w_{(u,w)} \epi(w)}{\gamma_{_0} (d(u)+\Delta d(u))^\beta d(w)^\nbeta}, \nonumber
\end{align}
where $\Delta d(u)$ is the weighted degree change of node $u$.


\section{Experimental Details}
To demonstrate the effectiveness of our proposed method, we compare the proposed method with competitive baselines on both CTDGs and DTDGs. Specifically, we perform future link prediction (self-supervised) and dynamic node classification tasks (semi-supervised) on seven real-world datasets. For baselines, we strictly inherit the performance of future link prediction reported in the TGN~\cite{rossi2020tgn} and ROLAND~\cite{you2022roland} papers, and the performance of each method in dynamic node classification reported in the TGL~\cite{zhou2022tgl} paper. We promise the same data partitioning and evaluation metric computation procedures as TGN~\cite{rossi2020tgn}, ROLAND~\cite{you2022roland} and TGL~\cite{zhou2022tgl} papers for fair comparisons. Our proposed method is implemented in PyTorch and C++. All experiments are performed on a machine outfitted with an NVIDIA RTX8000 GPU (48GB memory), an Intel Xeon CPU (2.20 GHz) with 40 cores, and 1TB of RAM.

\subsection{Datasets}
\begin{itemize}[leftmargin = *]
  \item \textbf{Wikipedia}~\cite{kumar2018jodie} and \textbf{Reddit}~\cite{kumar2018jodie} are bipartite graphs of interaction. In the Wikipedia dataset, users and web pages are nodes, while nodes in the Reddit dataset represent users and sub-reddits, and the label of each user indicates whether the user is banned or not. The edge $(i,j)$ indicates that user $i$ edits web page $j$ or posts on sub-reddit $j$, and the related feature vector represents the converted text features of the edit or post content.
  \item \textbf{UCI-MSG}~\cite{panzarasa2009uci} is an online user interaction graph based on social networks with nodes representing users. Each edge $(i, j)$ indicates that user $i$ sent a private message to user $j$.
  \item \textbf{Bitcoin-OTC}~\cite{kumar2016bitcoin, kumar2018bitcoin} and \textbf{Bitcoin-Alpha}~\cite{kumar2016bitcoin, kumar2018bitcoin} are user trust network based on the Bitcoin OTC and Bitcoin Alpha platform, respectively. Users are considered as nodes, and the edge $(i,j)$ indicates that user $i$ scores the extent of his trust to user $j$. This score ranges from -10 to 10, viewing as the one-dimensional edge feature. 
  \item \textbf{GDELT}~\cite{zhou2022tgl} is a temporal knowledge graph in which nodes represent participants, and CAMEO codes are utilized as node features. Each edge $(i, j)$ represents an event that occurs between participants $i$ and $j$. The participant is labeled with the country where the linked event occurred. Therefore, the node labels in GDELT change as time progresses.
  \item \textbf{MAG}~\cite{zhou2022tgl} is a large-scale dynamic paper citation network extracted from OGB-LSC~\cite{hu2021ogblsc}. In the MAG dataset, each node represents a paper, accompanied by a semantic vector produced from the paper abstracts and used as the node feature, and each paper is labeled with its arXiv subject area. The edge $(i, j)$ indicates that paper $i$ cites paper $j$.
\end{itemize}

\begin{table}[t]
\small
\caption{Hyperparameters for future link prediction.}
  \label{tab:parameter_link}
\vspace{-3mm}
\setlength\tabcolsep{3pt}
\begin{tabular}{cccccccc}
\hline
              & $\alpha$ & $\beta$ & $r_{max}$          & dropout & \begin{tabular}[c]{@{}c@{}}hidden\\ size\end{tabular} & \begin{tabular}[c]{@{}c@{}}batch\\ size\end{tabular} & \begin{tabular}[c]{@{}c@{}}learning\\ rate\end{tabular} \\ \hline
Wikipedia     & 0.2      & 0.5     & $1 \times 10^{-7}$ & 0.5     & 128                                                   & 128                                                  & 0.0001                                                  \\
Reddit        & 0.2      & 0.5     & $1 \times 10^{-7}$ & 0.5     & 128                                                   & 128                                                  & 0.0001                                                  \\
UCI-Message   & 0.2      & 0.5     & $1 \times 10^{-7}$ & 0.1     & 64                                                    & 1024                                                 & 0.001                                                   \\
Bitcoin-Alpha & 0.2      & 0.5     & $1 \times 10^{-7}$ & 0.1     & 128                                                   & 1024                                                 & 0.001                                                   \\
Bitcoin-OTC   & 0.2      & 0.5     & $1 \times 10^{-7}$ & 0.1     & 64                                                    & 1024                                                 & 0.001                                                  \\ \hline
\end{tabular}
\end{table}
\begin{table}[t]
\small
\caption{Hyperparameters for dynamic node classification.}
  \label{tab:parameter_node}
\vspace{-3mm}
\setlength\tabcolsep{3pt}
\begin{tabular}{cccccccc}
\hline
          & $\alpha$ & $\beta$ & $r_{max}$          & dropout & \begin{tabular}[c]{@{}c@{}}hidden\\ size\end{tabular} & \begin{tabular}[c]{@{}c@{}}batch\\ size\end{tabular} & \begin{tabular}[c]{@{}c@{}}learning\\ rate\end{tabular} \\ \hline
Wikipedia & 0.2      & 0.5     & $1 \times 10^{-7}$ & 0.1     & 128                                                   & 256                                                  & 0.0001                                                  \\
Reddit    & 0.2      & 0.5     & $1 \times 10^{-7}$ & 0.1     & 128                                                   & 256                                                  & 0.008                                                   \\
GDELT     & 0.2      & 0.5     & $1 \times 10^{-7}$ & 0.1     & 64                                                    & 2048                                                 & 0.001                                                   \\
MAG       & 0.2      & 0.5     & $5 \times 10^{-8}$ & 0.1     & 64                                                    & 2048                                                 & 0.001      \\ \hline
\end{tabular}
\end{table}
\subsection{Future Link Prediction}
{\revision
In the future link prediction task, the objective of model is to acquire the temporal patterns of node representation in dynamic graphs over time. To achieve this, we first complete the dynamic propagation and obtain the temporal representation of nodes as detailed in Section~\ref{sec:prediction_models}. We then employ commonly used sequence learning models for training and prediction. For the Wikipedia and Reddit datasets, we utilize a two-layer standard LSTM network to learn the temporal patterns of node representation. Specifically, we consider the temporal representation of each node as a sequence record, and the LSTM network predicts the representation of the node at the subsequent moment. For implementation, we use information prior to time $t$ to determine the state of the graph at time $t$. Assuming that the LSTM predicts the representations of nodes $i$ and $j$ as $h_{t,i}$ and $h_{t,j}$, respectively, for the edge $(i, j)$ at time $t$, we obtain its representation, $h_{t,(i,j)}$, by concatenating $h_{t,i}$ and $h_{t,j}$. The probability of the edge's existence at time $t$ is then determined by a two-layer MLP. Table~\ref{tab:parameter_link} summarizes the hyperparameter settings for the model. The propagation parameters include $\alpha$, $\beta$, and $r_{max}$, while the remaining hyperparameters are related to the neural network. We applied the same method used previously to obtain the probability of edge existence in the UCI-Message, Bitcoin-Alpha, and Bitcoin-OTC datasets. Additionally, we used the GRU cell and the four-head Transformer cell to replace the LSTM cell in order to demonstrate the effectiveness of different sequence models and to illustrate that our approach can integrate diverse sequence models seamlessly. Apart from the variation in sequence models, all other hyperparameters were kept constant across the three sets of experiments. Details of the specific hyperparameter settings can be found in Table~\ref{tab:parameter_link}.
}

\vspace{-2mm}
\subsection{Dynamic Node Classification}
{\revision
We evaluate the expressiveness of the generated temporal representation by conducting a dynamic node classification task. In the case of Wikipedia and Reddit, this task aims to identify unusual network users, which is modeled as a binary classification. To achieve this, we always use the most recent representation of a node and input it into a three-layer MLP. The output is the probability of the node belonging to a normal user. If the probability is less than 0.5, we classify the user as abnormal. The hyperparameter information for the experiment is listed in Table~\ref{tab:parameter_node}. The parameters are categorized into propagation parameters, which include $\alpha$, $\beta$ and $r_{max}$, as well as neural network hyperparameters such as dropout, hidden size, batch size, and learning rate. For GDELT and MAG, the experimental setup is similar, while the number of neurons in the output layer is adjusted to handle the multi-class classification task.
}

\section{Additional experimental results}
\subsection{Comparison on Time Cost}
\label{sec:comparison_time_cost}
{\revision 
Table~\ref{tab:exp_ctdg_timecost} is a comparison of the time needed by our method and baselines to complete a single training epoch. For TGAT~\cite{xu2020tgat} and CAW-Ns~\cite{wang2021caw}, we utilize their official implementations and default hyper-parameters. The implementation of JODIE~\cite{kumar2018jodie}, DyRep~\cite{trivedi2019dyrep} and TGN~\cite{rossi2020tgn} follows Rossi et al~\cite{rossi2020tgn}. We found that compared to baselines, our method requires much less time. Since the propagation phase previously completed the time-consuming computations related to the graph topology, and the training process optimizes only the sequence model's parameters.
}
\begin{table}[t]
\caption{\revision Comparison of the computation time for single-epoch of training (second).}
\label{tab:exp_ctdg_timecost}
\begin{tabular}{c|c|c}
\hline
         & Wikipedia            & Reddit   \\ \hline
JODIE    & 10 ± 0.05            & 78 ± 0.98      \\
TGAT     & 147 ± 5.97           & 764 ± 1.98      \\
DyRep    & 12 ± 0.64            & 93 ± 3.21       \\
TGN      & 12 ± 0.12            & 90 ± 1.22       \\
CAW-N-mean & 388 ± 1.62         & 2046 ± 102.34     \\
CAW-N-attn & 398 ± 1.14         & 2010 ± 60.47      \\
ours     & 2 ± 0.32             & 12 ± 0.87       \\ \hline               
\end{tabular}
\end{table}
\begin{table*}[t]
\caption{Ablation study on the effectiveness of high-frequency information. MRR $\pm$ standard deviations computed of 3 random seeds are exhibited.}
  \label{tab:exp_ablation_study}
\vspace{-3mm}
\begin{tabular}{cc|cccccc}
\hline
               &                              & UCI-Message     & gain                    & Bitcoin-Alpha   & gain                     & Bitcoin-OTC     & gain                    \\ \hline
ours-w.o.-high & \multirow{2}{*}{GRU}         & 0.1962 ± 0.0073 & \multirow{2}{*}{3.16\%} & 0.3270 ± 0.0055 & \multirow{2}{*}{0.58\%}  & 0.2943 ± 0.0121 & \multirow{2}{*}{1.43\%} \\
ours-concat    &                              & 0.2024 ± 0.0010 &                         & 0.3289 ± 0.0070 &                          & 0.2985 ± 0.0121 &                         \\ \hline
ours-w.o.-high & \multirow{2}{*}{LSTM}        & 0.2127 ± 0.0084 & \multirow{2}{*}{0.61\%} & 0.3163 ± 0.0046 & \multirow{2}{*}{7.65\%}  & 0.3054 ± 0.0048 & \multirow{2}{*}{1.57\%} \\
ours-concat    &                              & 0.2140 ± 0.0034 &                         & 0.3405 ± 0.0133 &                          & 0.3102 ± 0.0046 &                         \\ \hline
ours-w.o.-high & \multirow{2}{*}{Transformer} & 0.2249 ± 0.0018 & \multirow{2}{*}{2.89\%} & 0.3290 ± 0.0124 & \multirow{2}{*}{-3.56\%} & 0.3035 ± 0.0035 & \multirow{2}{*}{2.47\%} \\
ours-concat    &                              & 0.2314 ± 0.0048 &                         & 0.3173 ± 0.0135 &                          & 0.3110 ± 0.0049 &                           \\ \hline
\end{tabular}
\end{table*}

\vspace{-2mm}
\subsection{Ablation Study}
\label{sec:ablation_study}
In this subsection, we conduct an ablation study to verify the effectiveness of high-frequency information, which is concatenated with low-frequency information and used simultaneously in the future link prediction task. We design two variants for the high-frequency information concatenation strategies as follows: 
\begin{itemize}[leftmargin = *]
  \item \textbf{ours-w.o.-high}: we remove the high-frequency information generated by $\gamma_k=\alpha(\alpha-1)^k$.
  \item \textbf{ours-concat}: we concatenate the low- and high-frequency information, which is used in Section~\ref{sec:exp_dtdg}.
\end{itemize}
We conduct the ablation study on Bitcoin-OTC, Botcoin-Alpha and UCI-Message datasets. The experimental results are shown in Table~\ref{tab:exp_ablation_study}. In most cases, ours-concat performs better than ours-w.o.-high, indicating that the inclusion of high-frequency information improves the expressiveness of the model.

\subsection{Parameter Sensitivity}
\label{sec:parameter_sensitivity}
\begin{figure}[t]
\setlength{\abovecaptionskip}{1mm}
\setlength{\belowcaptionskip}{-2mm}
\begin{small}
    \centering
    \vspace{-2mm}
    \begin{tabular}{cc}
         \hspace{-3mm}\includegraphics[height=25mm]{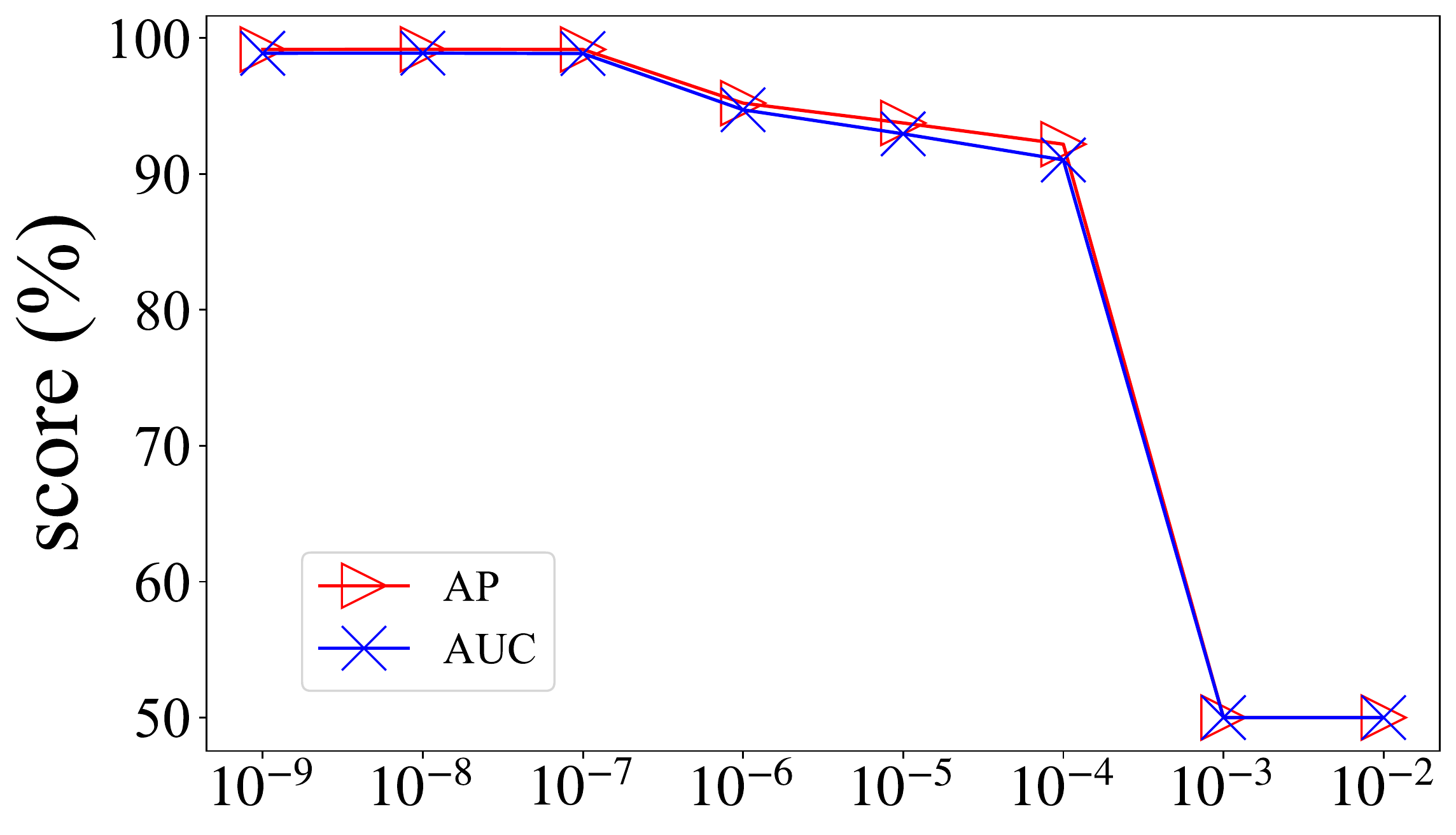} &
         \hspace{-3.5mm}\includegraphics[height=25mm]{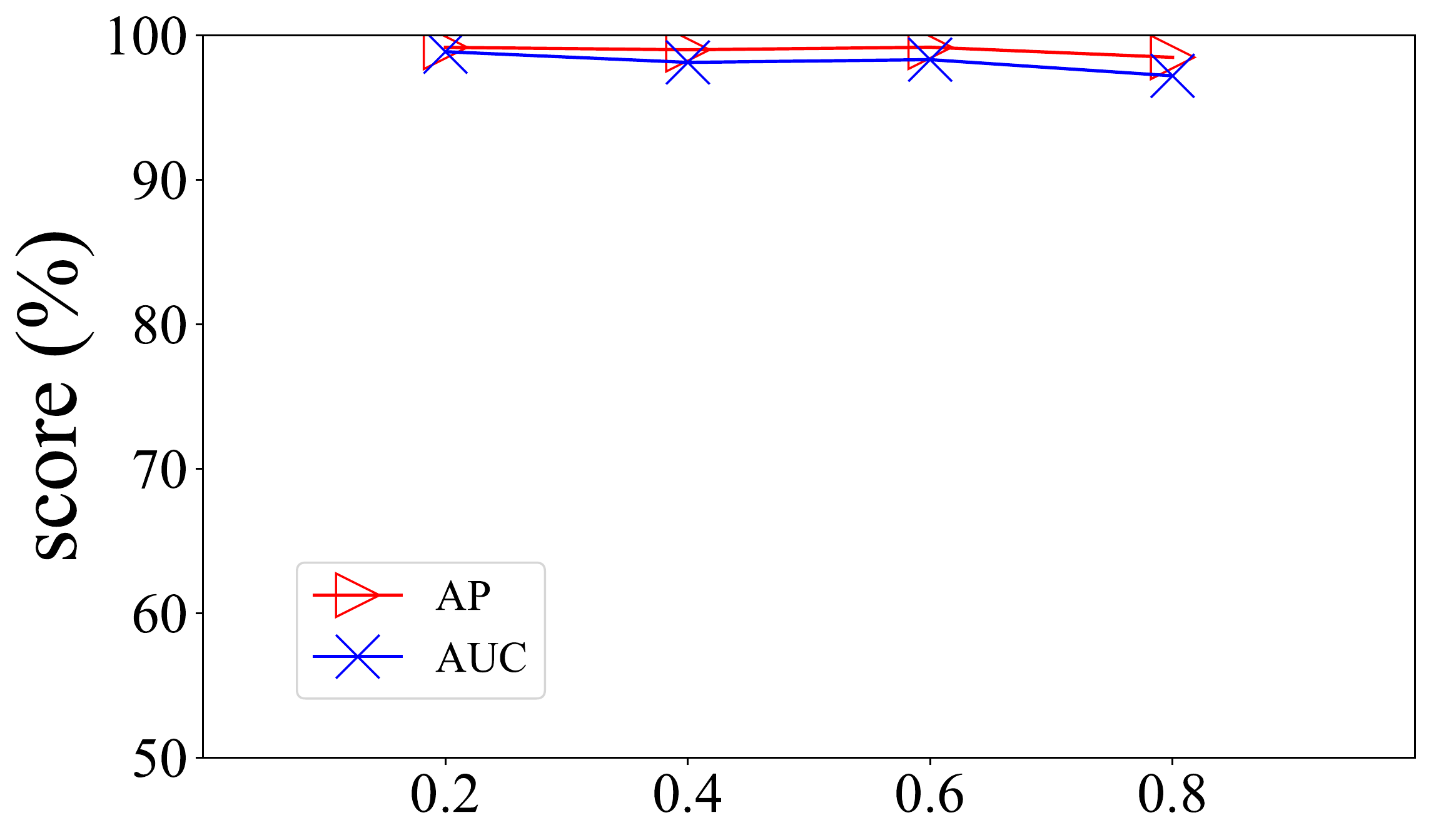} \\ 
            (a) The threshold $r_{max}$. & (b) The weight coefficient $\alpha$. \\
         \hspace{-3mm}\includegraphics[height=25mm]{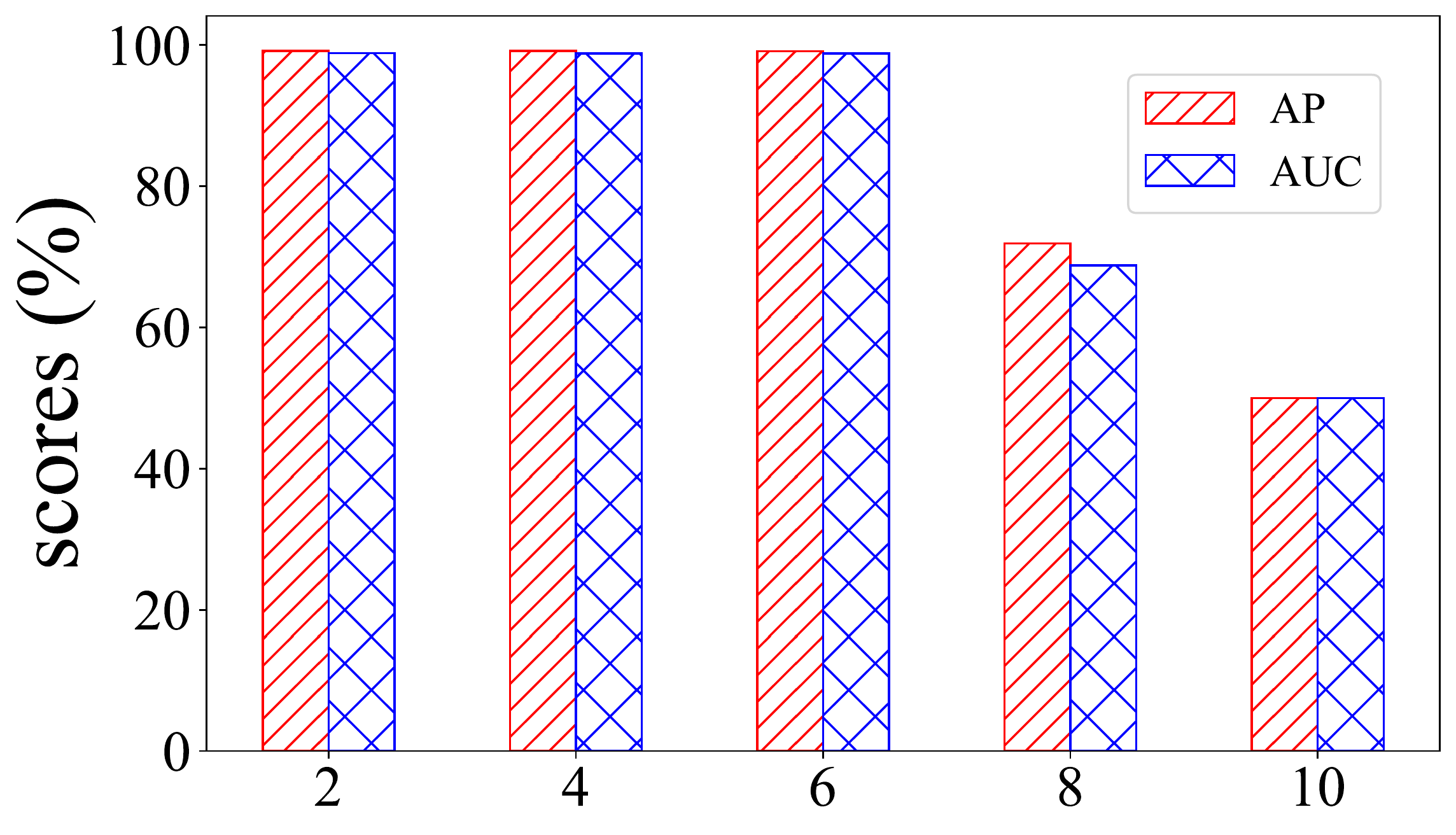} &
         \hspace{-3.5mm}\includegraphics[height=25mm]{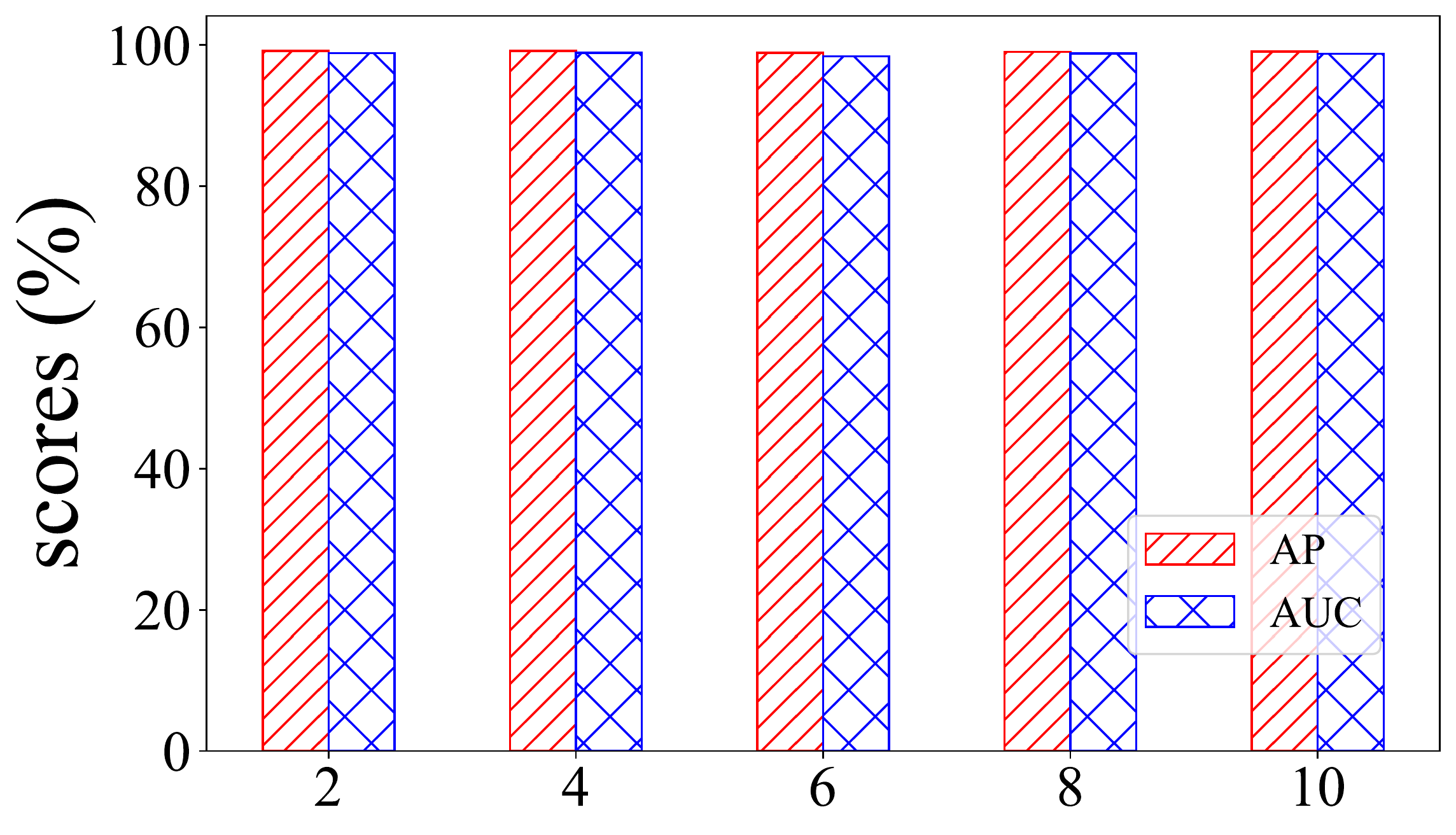} \\
            (c) The number of LSTM layers. & (d) The number of MLP layers.
    \end{tabular}
    \caption{\revision Parameter sensitivity on Wikipedia dataset.}
    \label{fig:params_sensitivity}
\end{small}
\end{figure}

{\revision
We group the parameters into the propagation and the neural network parameters in accordance with the decoupled architecture. According to Algorithms~\ref{alg:static}\&~\ref{alg:dynamic}, the propagation parameters include $r_{max}$ and $\gamma$, where $r_{max}$ is directly connected to the embedding's quality as mentioned in Section~\ref{sec:error_analysis}. We first examine the effectiveness of $r_{max}$. Specifically, we run 8 experiments with $r_{max} \in \{10^{-9}, 10^{-8}, …, 10^{-2}\}$ to study the effect of $r_{max}$ on performance. Here, $r_{max} = 10^{-9}$ means that the node embedding is sensitive to changes happening both inside itself and in its neighborhood, while $r_{max} = 10^{-2}$ indicates that the embedding has a high error tolerance and is hence not sensitive to changes. Figure~\ref{fig:params_sensitivity}(a) illustrates the test performance of our algorithm on the Wikipedia dataset with various $r_{max}$ values. As previously stated, a lower $r_{max}$ number often results in improved model performance and reduced error in node embedding. Nevertheless, this may also cause an increase in propagation time or cause the neural network's parameter-sensitive period to be exceeded. In addition, $r_{max}$ values of $10^{-3}$ and $10^{-2}$ render the model incorrect since the allowed error of node embedding is too great, making it impossible to have graph events that may cause it to change. Therefore, $r_{max}$ is often set to $10^{-7}$ in most of the datasets to balance model accuracy and computation time. According to section~\ref{sec:error_analysis}, we set $\gamma_k = \alpha (1-\alpha)^k$ to eliminate the influence of $\gamma$ setting on the accuracy of node embedding. As shown in Figure~\ref{fig:params_sensitivity}(b), the model performance is scarcely affected by adjustments of $\alpha$, which meets our requirement. Figure~\ref{fig:params_sensitivity}(c) and Figure~\ref{fig:params_sensitivity}(d) illustrate the influence of the number of LSTM and MLP layers, respectively. We observe that increasing the number of layers has little effect on the performance of the model. However, when the number of LSTM layers surpasses eight, the model has too many parameters, making convergence problematic and resulting in poor performance.
}

\section{Time encoding}
\label{sec:time_encoding}
{\revision
TGAT~\cite{xu2020tgat} employs a time encoding function to encode the relative time difference between nodes' history interactions and the present, and subsequent works such as TGN~\cite{rossi2020tgn} follow this strategy to assign weights to nodes' historical neighbors in chronological order. The underlying premise of these methods is that the embedding of a node only changes when it interacts. In our decoupled architecture, the change of a node's embedding is controlled by the threshold $r_{max}$ and could be caused by its associated events or the change in its neighborhood. There are two possible scenarios: (1) When $r_{max}$ is set to a larger number, such as 0.1, and node $u$ has a new or deleted edge at time $t$, but the impact produced by the edge is insignificant, i.e., the residual at node $u$ does not exceed the error tolerance, we consider that node $u$'s embedding is unchanged at that time. (2) When $r_{max}$ is set to a lesser number, such as $10^{-10}$, node $u$ has no new or deleted edges at time $t$, but its neighboring nodes have changed dramatically, resulting in a change in its embedding. Note that we address the preceding two cases in the exact opposite way as TGAT~\cite{xu2020tgat} and TGN~\cite{rossi2020tgn}, which update the embedding of node $u$ in the first case but not the second. 

We decouple the propagation from the training process so that computations pertaining to the graph structure can be performed beforehand. This phase focuses on providing a temporal representation for each node, which could be regarded as a description of the evolution of graph from each node's perspective, under the supervision of a carefully designed propagation formula. \eat{Therefore, once the graph has changed, each node $u$ has corresponding feedback indicating the degree of the change impacts it. } As a result, if the graph changes, each node $u$ responds by signaling how much the change has an impact on it. Consequently, the final temporal sequence we construct for each node already has the information regarding the time difference between embedding modifications. Furthermore, time encoding can be naturally introduced as supplemental information to the temporal representation in our method. That is, each feedback from node $u$ may correlate to quantified time interval information. Table~\ref{tab:exp_time_encoding} illustrates the results of our method with the addition of time encoding, where the results are all obtained from the LSTM model. In our implementation, the time encoding is combined with the temporal representation generated from propagation and put into the LSTM model to learn the dynamic graph's temporal pattern. As is shown in Table~\ref{tab:exp_time_encoding}, the addition of time encoding resulted in a certain degree of model improvement, but the improvement is not leapfrog and corresponds to our earlier discussion.
}
\begin{table}[t]
\caption{\revision Effect of time-encoding function ({\em TE}).}
  \label{tab:exp_time_encoding}
\vspace{-3mm}
\setlength\tabcolsep{3pt}
\begin{tabular}{l|c|ccc}
\hline
    & AP        & \multicolumn{3}{c}{MRR} \\ \hline
    & Wikipedia & UCI-Message & Bitcoin-Alpha & Bitcoin-OTC \\ \hline
w.o. {\em TE} & 99.16     & 0.214       & 0.3405        & 0.3102      \\
w. {\em TE}   & 99.20     & 0.215       & 0.3423       & 0.3208      \\ \hline
\end{tabular}
\end{table}

\end{document}